\documentclass[10pt,journal,compsoc]{IEEEtran}
\ifCLASSOPTIONcompsoc
  \usepackage{cite}
\else
  \usepackage{cite}
\fi
\usepackage{times}
\usepackage{epsfig}
\usepackage{graphicx}
\usepackage{amsmath}
\usepackage{amssymb}
\usepackage{url}
\usepackage{algorithm}
\usepackage{algorithmicx}
\usepackage{enumitem}
\usepackage{bm}
\usepackage[noend]{algpseudocode}
\usepackage[table, dvipsnames]{xcolor}
\usepackage{relsize}
\usepackage{exscale}

\newcommand{\red}[1]{\textcolor{black}{#1}}

\DeclareMathOperator*{\argmax}{arg\,max}
\DeclareMathOperator*{\argmin}{arg\,min}

\newcommand\descitem[1]{\item{\bfseries #1}}
\newcommand{\norm}[1]{\left\lVert#1\right\rVert}
\usepackage{comment}
\usepackage{amsfonts}

\begin{document}
\title{Explaining Deep Face Algorithms through Visualization: A Survey}

\author{Thrupthi~Ann~John,
        Vineeth~N~Balasubramanian,~\IEEEmembership{Senior Member,~IEEE,}
        and~C~V~Jawahar,~\IEEEmembership{Member,~IEEE}
\IEEEcompsocitemizethanks{\IEEEcompsocthanksitem Thrupthi Ann John  is and C V Jawahar are with the International Institute of Information Technology, Hyderabad \protect\\
E-mail: thrupthi.ann@research.iiit.ac.in 
\IEEEcompsocthanksitem Vineeth N Balasubramanian is with the Indian Institute of Technology, Hyderabad. \protect\\

}
}

\markboth{IEEE Transactions on Biometrics, Behavior, and Identity Science (T-BIOM)}%
{John \MakeLowercase{\textit{et al.}}:Explaining Deep Face Algorithms through Visualization}

\IEEEtitleabstractindextext{%
\begin{abstract}

Although current deep models for face tasks surpass human performance on some benchmarks, we do not understand how they work. Thus, we cannot predict how it will react to novel inputs, resulting in catastrophic failures and unwanted biases in the algorithms. Explainable AI helps bridge the gap, but currently, there are very few visualization algorithms designed for faces. This work undertakes a first-of-its-kind meta-analysis of explainability algorithms in the face domain. We explore the nuances and caveats of adapting general-purpose visualization algorithms to the face domain, illustrated by computing visualizations on popular face models. We review existing face explainability works and reveal valuable insights into the structure and hierarchy of face networks. We also determine the design considerations for practical face visualizations accessible to AI practitioners by conducting a user study on the utility of various explainability algorithms. 

\end{abstract}

}

\maketitle

\IEEEdisplaynontitleabstractindextext

\IEEEpeerreviewmaketitle

\ifCLASSOPTIONcompsoc

\newpage

\section{Introduction}
\label{sec:introduction}
Deep Neural Network (DNN) models have recently shown unprecedented progress for various face processing tasks like face recognition, emotion recognition, head pose estimation, age and gender recognition -- often surpassing human benchmarks. This rapid growth in the face domain using deep learning models led to its use in safety-critical applications, including fraud avoidance in rental cars by face verification, identifying hallucinated faces and detecting the driver’s head pose to avoid accidents on the road. Models for face tasks are already available from major businesses like Microsoft, IBM, and Amazon, all of which claim to be very accurate. However, a limited understanding of how these models work restricts their applicability and trustworthiness.

\begin{figure}
	\centering
	\includegraphics[width=0.85\linewidth]{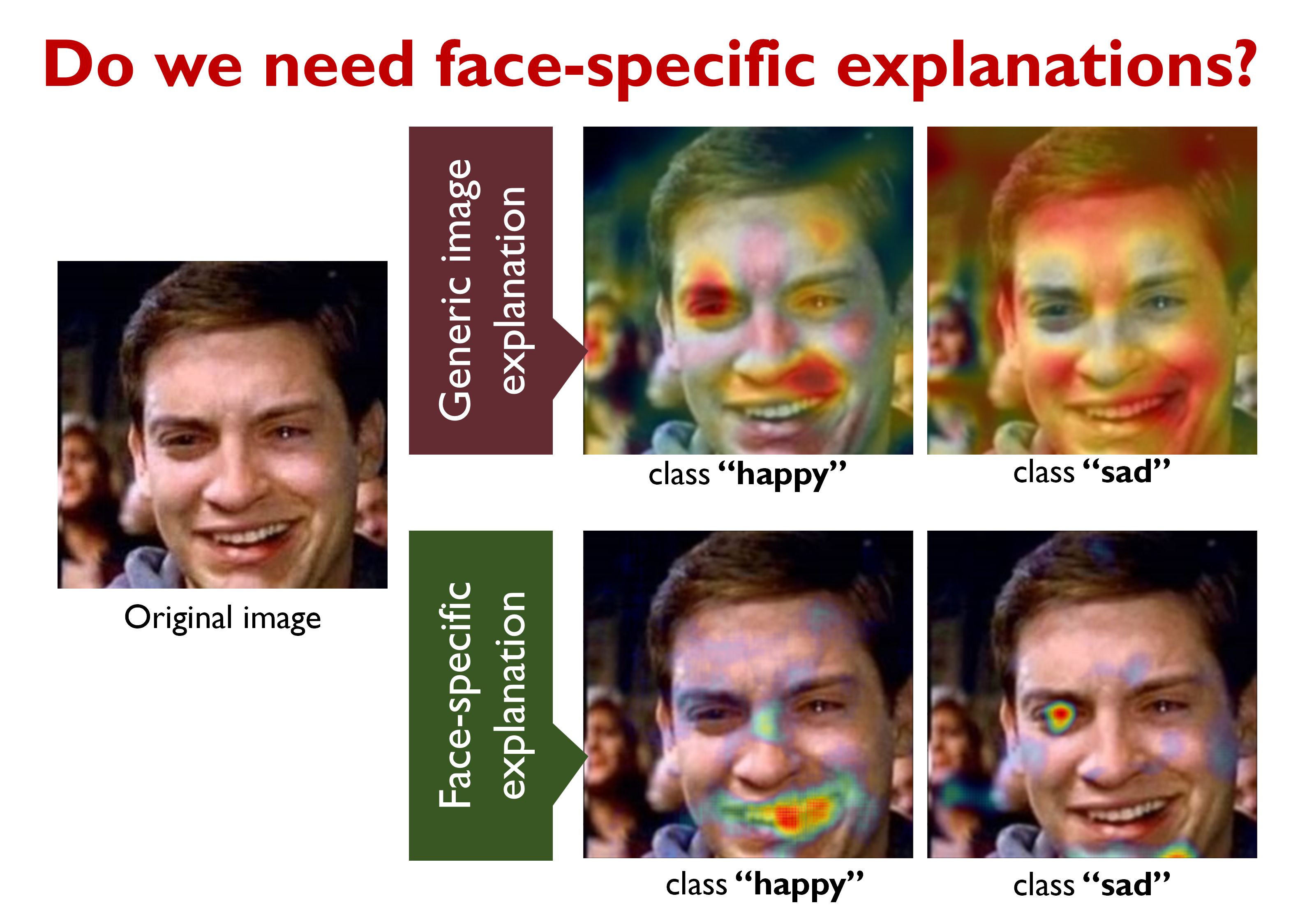}
	\vspace{-6pt}
	\caption{Although there is extensive research on general explainability methods in recent years, many general algorithms do not work well on faces directly without explicit modification \cite{john2021canonical, zhong2018difference}. Pictured above is the comparison of applying a generic model explanation method, GradCAM \cite{selvaraju2017gradcam} \textit{(top)}, against Occlusion \cite{zeiler2014visualizing}, a method known to work on faces (bottom). In the original image, the face seems to have conflicting emotions: the lips are `smiling' while the eyes are `crying'. GradCAM fails to highlight relevant parts for each class, whereas Occlusion produces a better explanation for each class.}
	\label{fig:titlefig}
\end{figure}

There is a growing recognition of the importance and relevance of explainability across Artificial Intelligence (AI) applications. Critical applications such as automated driving and financial decision-making systems need to be highly reliable, as any errors may be catastrophic. Recent works have shown how self-driving cars are susceptible to adversarial attacks \cite{chernikova2019self, eykholt2018robust}. Explainability improves the systems' trustworthiness and allows the users to evaluate potential failure modes and pinpoint the causes of such failures. Many critical AI applications are protected by ethical and legal requirements that the systems should avoid algorithmic discrimination, are unbiased/fair and can explain their decision process. Algorithmic drug discovery \cite{gawehn2016drug}, for another example, needs to be interpretable, as the US Food and Drug Administration (FDA) requires an explanation of the biological mechanisms of a result. The Right to Explanation \cite{goodman2017european} guarantees an individual's right to be given an explanation for decisions that significantly affect the individual, particularly legally or financially. According to the EU General Data Protection Regulation (GDPR) \cite{regulation2016regulation} Article 22, people have the right not to be subject to an automated decision which would produce legal effects concerning them. We cannot guarantee these rights without proper explanations of our algorithms. 

While many explainability surveys exist \cite{zhang2021survey, shahroudnejad2021survey, samek2021survey, das2020opportunities, carvalho2019machine, buhrmester2021analysis, guidottiSurveyMethodsExplaining2018}, surprisingly, there is no survey or meta-analysis of explainability for deep face models, despite the widespread use of face-based DNN models. Most current explainability methods are designed for object recognition and cannot directly be applied to the face domain due to their differences from the natural image domain, as shown in Figure \ref{fig:titlefig}. Unlike the wide variety of objects, faces are highly structured, and the variations in features, colors and shapes differ from other domains. Tasks in the face domain are often fine-grained. Some popular face tasks are inconsistently defined, and the relationship between facial features and tasks is subtle. For example, a smile may not always indicate joy, and a frown may be due to surprise, anger or disapproval. There may be no direct mapping between wrinkles and age. Some attributes, such as criminal propensity or intelligence, do not correlate with facial features. We need to be sensitive to these issues while processing or generating faces. There are also dependencies between face tasks. For example, a man’s smile (in an expression recognition task) may differ from a woman’s due to structural differences (gender recognition task), or hair color may depend on a person’s age. There is hence a need to understand and process faces holistically, with specialized algorithms for explaining face models. 

This work aims to bridge the gap in explainability literature for the face domain. We present promising directions for the interpretability of deep neural network models, with a focus on face processing models. We show the results of their application to various face processing tasks and explore their caveats. We also share deep insights on DNN-based face processing models obtained using these explainability methods and examine their relation to human understanding of faces.

\begin{table*}
	\centering
	{\rowcolors{1}{lightgray!80!white!50}{white}
		\begin{tabular}{|p{7.5cm}|c|c|c|c|}
			\hline
			Name & Authors & Year & Type & Use \\
			\hline
			Canonical Saliency Maps: Decoding Deep Face Models \cite{john2021canonical} &  John et al. & 2021 & Image and model saliency & Interpreting models \\
			Explainable Face Recognition \cite{williford2020explainable} & Williford et al. & 2020 & Image saliency & Interpreting models \\
			Exploring Features and Attributes in Deep Face Recognition Using Visualization Techniques \cite{Zhong2019Exploring} & Zhong \& Deng & 2019 & Various & Interpreting models\\
			Enhancing Human Face Recognition with an Interpretable Neural Network \cite{zee2019enhancing} & Zee et al. & 2019 & Various & Assisting humans \\
			Towards Interpretable Face Recognition \cite{yin2019Towards} & Yin et al. & 2019 & Intrinsic explainability & Interpreting models \\
			Deep Difference Analysis in Similar-looking Face recognition \cite{zhong2018difference} & Zhong \& Deng & 2018 & Image saliency & Assisting humans\\
			Visual Psychophysics for Making Face Recognition Algorithms More Explainable \cite{richardwebster2018psychophysics} & RichardWebster et al. & 2018 & Evaluation for explainability & Interpreting models \\
			Visualizing and Quantifying Discriminative Features for Face Recognition \cite{castanon2018Visualizing} & Castanon \& Byrne & 2018 & Image saliency & Interpreting models \\
			Learning to Predict Saliency on Face Images \cite{Xu2015LearningGaze} & Xu et al. & 2015 & Human visual saliency  & -\\
			\hline
	\end{tabular}}
	\caption{Summary of explainability papers on the face domain categorized according to the classification in Section \ref{sec:introduction}.}
	\label{tab:facepapers}
\end{table*}

\vspace{6pt}
\noindent \textbf{Classification of Explainability Methods.}
Before we state our contributions on assessing visual explanation methods for DNN-based face processing models, we briefly review the general literature on explainability methods. Existing methods for explainability have been categorized in recent literature surveys \cite{buhrmester2021analysis, carvalho2019machine, das2020opportunities, shahroudnejad2021survey, zhang2021survey}. In this work, we classify the methods along four dimensions: \textbf{Why}, \textbf{What}, \textbf{When} and \textbf{Scope}, as shown in Figure \ref{fig:classification}..

The first dimension, ``\textit{Why}'', explores the reason for applying the explanation algorithm. General explanation algorithms are usually applied to explain the working of a DNN model. For example, Section \ref{sec:saliency} discusses using saliency visualization to understand biases in some face processing models. However, in the face domain, explanation algorithms are also used to assist humans in determining the correctness of a model's decisions or making better judgements in face tasks. We discuss some examples in Section \ref{sec:saliencyinsights}. 

\begin{figure}
	\centering
	\includegraphics[width = 0.9\linewidth]{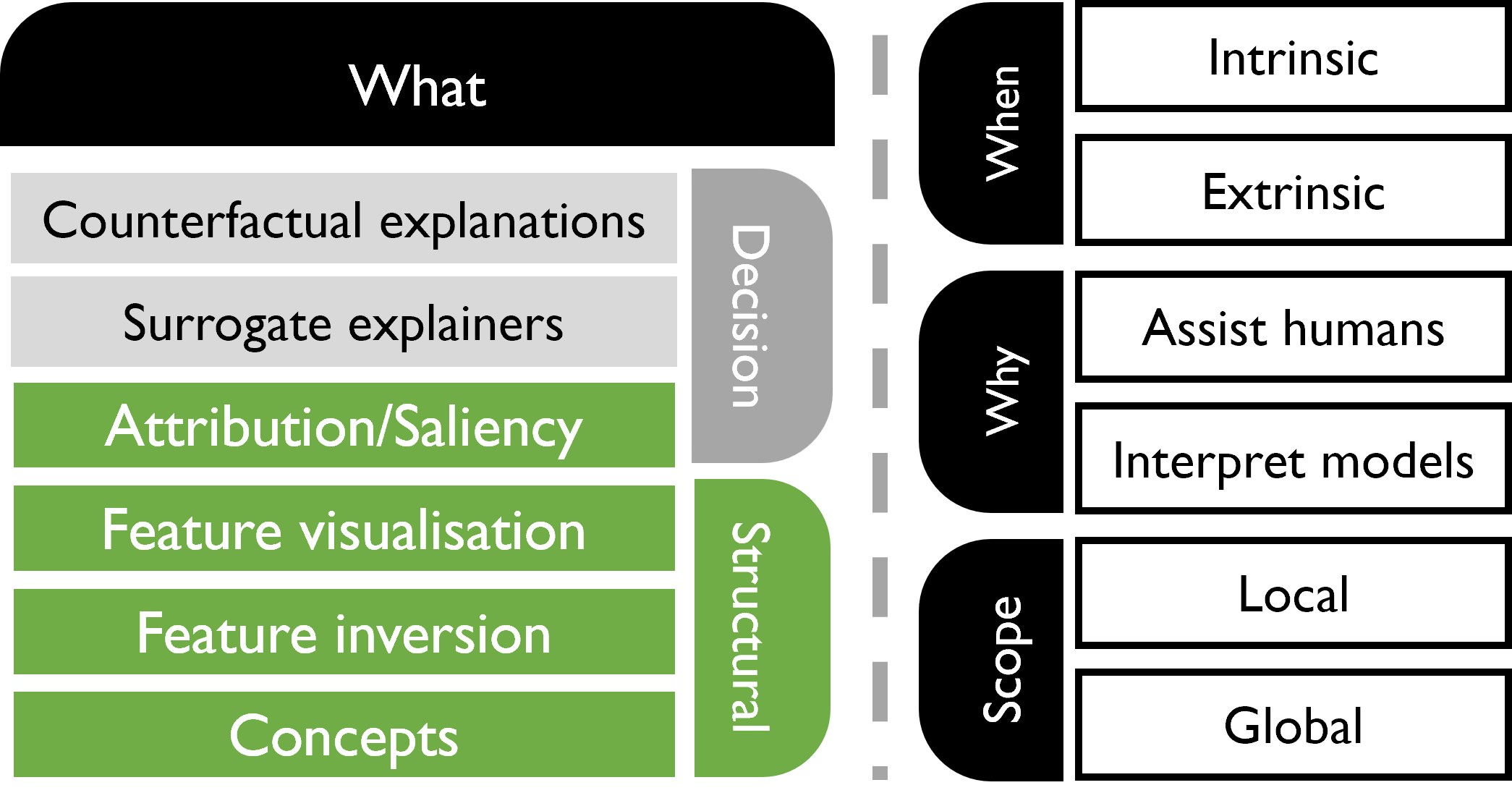}
	\vspace{5pt}
	\caption{We categorize explainability methods across four dimensions: \textbf{Why} represents the end goal of applying the algorithm; \textbf{What} describes the part of the model we are explaining; \textbf{When} describes if the algorithm is applied during or after training; \textbf{Scope} describes if the algorithm explains a single input or the entire model. The \textbf{What} component can be divided into \textit{decision analysis} and \textit{structural interpretation}. We focus on the latter aspects in this work.}
	\label{fig:classification}
\end{figure}

In the ``\textit{What}'' dimension, based on which model components we are explaining, explainability methods can be divided into \emph{decision analysis} -- that explore what factors go into the predictions, and \emph{structural interpretation} -- that visualize the internal representations and learned weights. Under decision analysis, attribution analysis determines which parts of an input highly influence the outcome using saliency heatmaps. Surrogate explainers try to reduce a DNN model to a naturally interpretable AI model like a decision tree, rule-based model or linear classifier. Counterfactual explanations describe the smallest change to the feature values that changes the prediction to a predefined output. We do not cover surrogate explainers and counterfactual explanations in this work as they are not visual explainability methods per se. We primarily discuss two types of structural visualizations in this work: (i) \emph{Feature visualization} methods visualize the internal representations of different components of a DNN model projected into the input domain; (ii) \emph{Feature inversion} methods take the internal representation corresponding to an input image and project it back to the input domain. Often, these visualizations may not correspond to recognizable human abstractions and constructs. We also discuss a third kind, \emph{Concept-based explanation} methods, that attempt to align the internal features and representations with human concepts in Section \ref{sec:concepts}.

The ``\textit{When}'' category of methods refers to which part of the training/deployment process the explainability algorithm is applied. Models with intrinsic explainability have interpretability baked into their structure. The network architecture, losses and training procedure need to be changed; hence, they may not be model-agnostic. Intrinsic explainability in deep models means that each component/neuron of the model represents a single or limited type of well-defined object part. Enforcing this often leads to reduced performance. On the other hand, post-hoc methods consider the model a black box and are applied to pre-trained models. They are primarily model-agnostic. They are generally more flexible than intrinsic methods as they do not require modification to the training procedure. However, they tend to be approximate methods as they guess at the 'intention' of the models. Finally, the ``\textit{Scope}'' dimension refers to whether the explainability algorithm explains the model for a few inputs (\textit{local}) or the entire model at once (\textit{global}). We now discuss our contributions.

\vspace{6pt}
\noindent \textbf{Our Contributions.}
As stated earlier, while there have been earlier efforts on reviewing existing work on machine learning explainability \cite{zhang2021survey, shahroudnejad2021survey, samek2021survey, das2020opportunities, carvalho2019machine, buhrmester2021analysis, guidottiSurveyMethodsExplaining2018}, ours is the first to focus on a meta-analysis of face explainability. This survey will be helpful to researchers looking to understand the structure of face models and to apply explainability algorithms to face models to increase their interpretability. 

Our contributions are as follows:
\begin{enumerate}
	\item We present an overview of explainability methods, especially visualization/structural explainability methods, and show their application to the face domain. We describe the issues that arise from directly applying general explainability methods to faces and how to modify the algorithms to work on face models. We also  discuss the insights we gain from them on the workings of face models. 
	\item We review explainability literature specifically targeted to face processing models. To our knowledge, we are the first work to survey face explainability works. We emphasize on the need for face-specific methods of evaluation for explainability. 
	\item We identify factors required to make face explainability methods more accessible to AI practitioners by conducting a user survey on the utility of different kinds of explainability algorithms.  
\end{enumerate}

The remainder of this paper is organized as follows. Sections \ref{sec:saliency}, \ref{sec:featurevisualisation} and \ref{sec:featureinversion} present an overview of different well-known post-hoc visualization methods and their application to the face domain. Section \ref{sec:concepts} presents a discussion on how to align features to human concepts. Section \ref{sec:faceevaluation} subsequently describes evaluation protocols for face algorithms inspired by psychophysics. Finally, in Section \ref{sec:usersurvey}, we report a user study on the utility and human-interpretability of well-known visualization methods for face processing models.

\section{Saliency Maps}
\label{sec:saliency}

\begin{figure}
	\centering
	\includegraphics[width=0.7\linewidth]{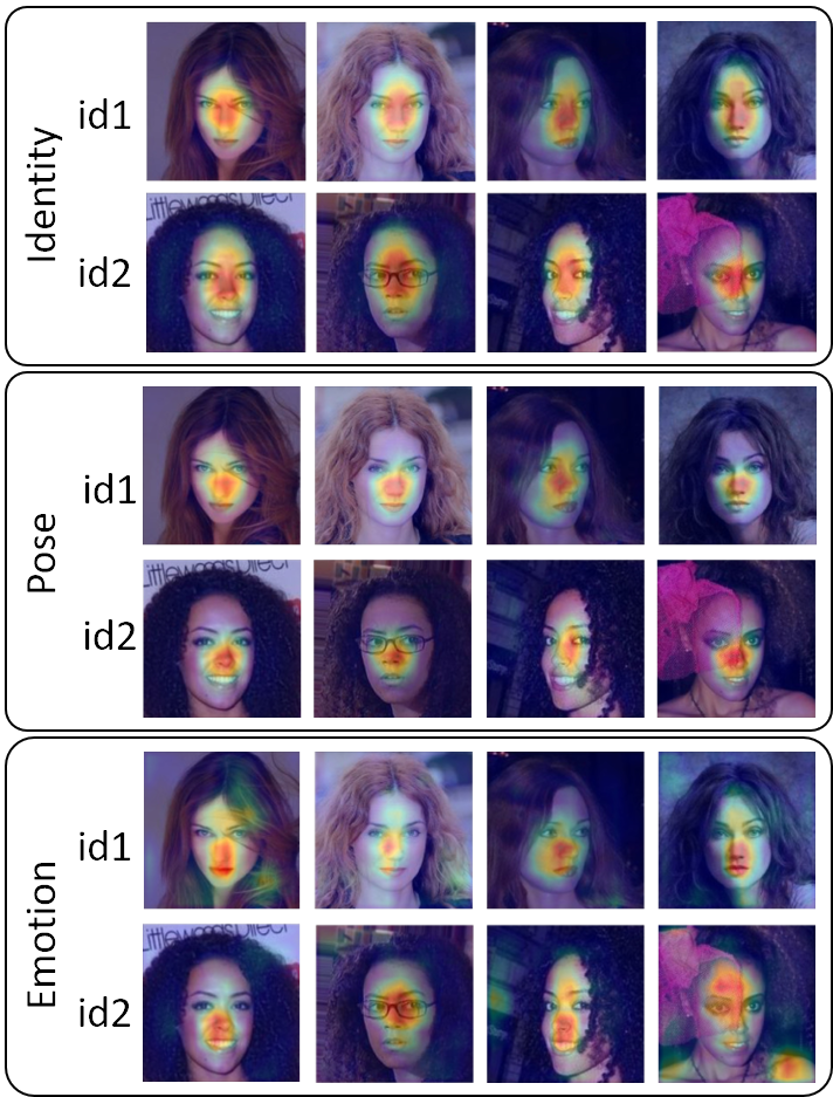}
	\vspace{-4pt}
	\caption{Saliency algorithms designed for general object recognition often do not work well on faces. Here we see GradCAM \cite{selvaraju2017gradcam} applied to three models trained on different face tasks. Despite differences in identity, expression and pose, we see that the algorithm focuses generally on the center of the face. Original images are taken from the VGG-Face dataset \cite{parkhi2015vggface}. See also Figure \ref{fig:titlefig}.}
	\label{fig:face_gradcam}
\end{figure}

\begin{figure*}
	\centering
	\includegraphics[width=0.95\linewidth]{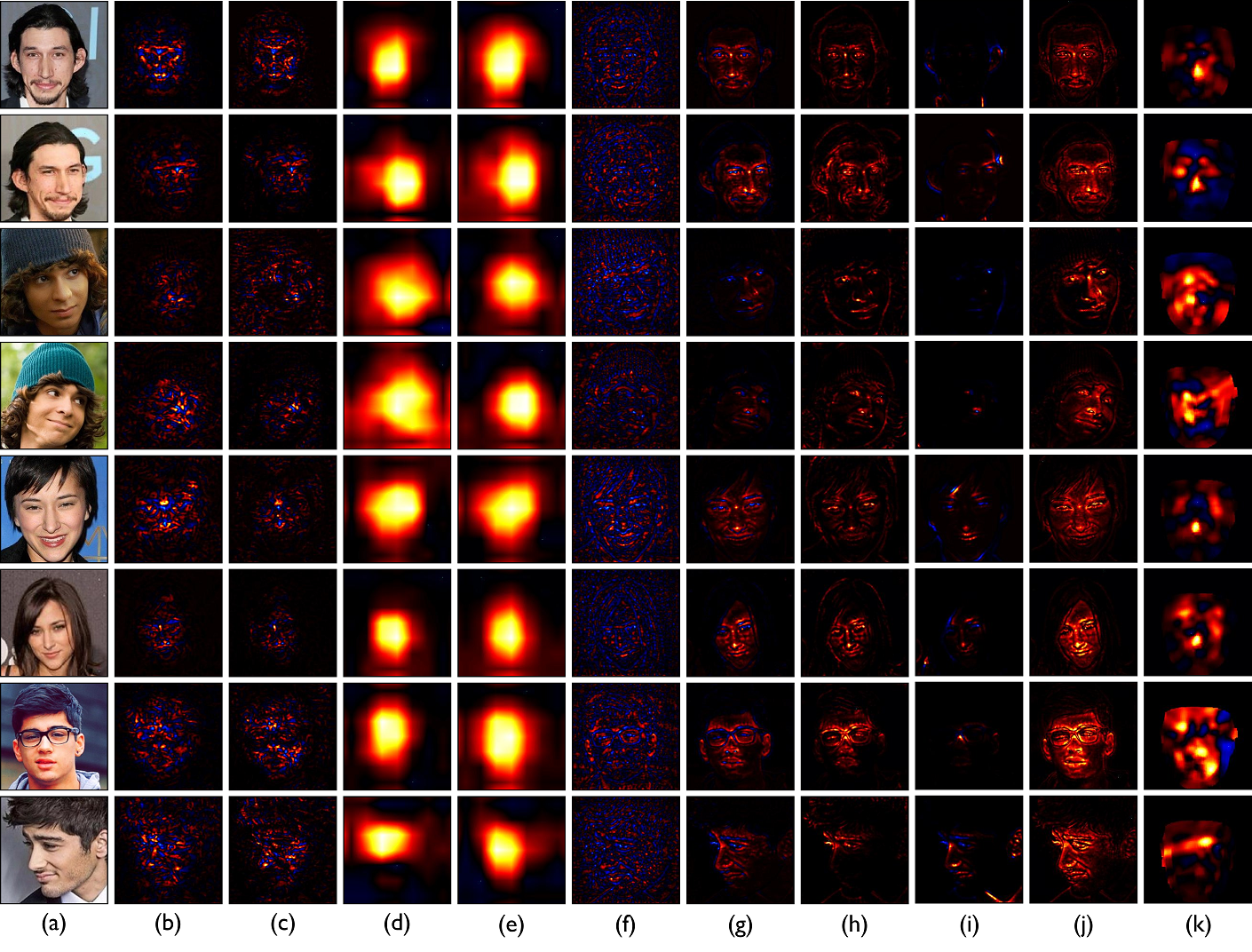}
	\vspace{-9pt}
	\caption{Comparison of various saliency visualization methods on the VGG-Face model \cite{parkhi2015vggface} for the task of face recognition. For each image, the target class of the visualization is the ground truth class. Red to yellow represents increasing positive saliency, blue to cyan represents increasing negative saliency and black stands for neutral. (a) Original image; (b) Smoothgrad \cite{smilkov2017smoothgrad}; (c) Integrated gradients \cite{sundararajan2017axiomatic}; (d) GradCAM \cite{selvaraju2017gradcam}; (e) ScoreCAM \cite{wang2020scorecam}; (f) Deconvolution \cite{zeiler2014visualizing}; (g) Guided Backpropagation \cite{selvaraju2017gradcam}; (h) LRP \cite{bach2015LRP}; (i) DeepLIFT \cite{shrikumar2017deeplift}; (j) Excitation Backprop \cite{zhang2018excitationbackprop}; (k) Canonical Saliency maps \cite{john2021canonical}. The original images are taken from the VGG-Face dataset \cite{parkhi2015vggface}. Rows (1, 2), (3, 4), (5, 6) and (7, 8) have the same identity. Original images are taken from the VGG-Face dataset \cite{parkhi2015vggface}.}
	\label{fig:face_attention_survey}
\end{figure*}

\begin{figure}
	\centering
	\includegraphics[width=\linewidth]{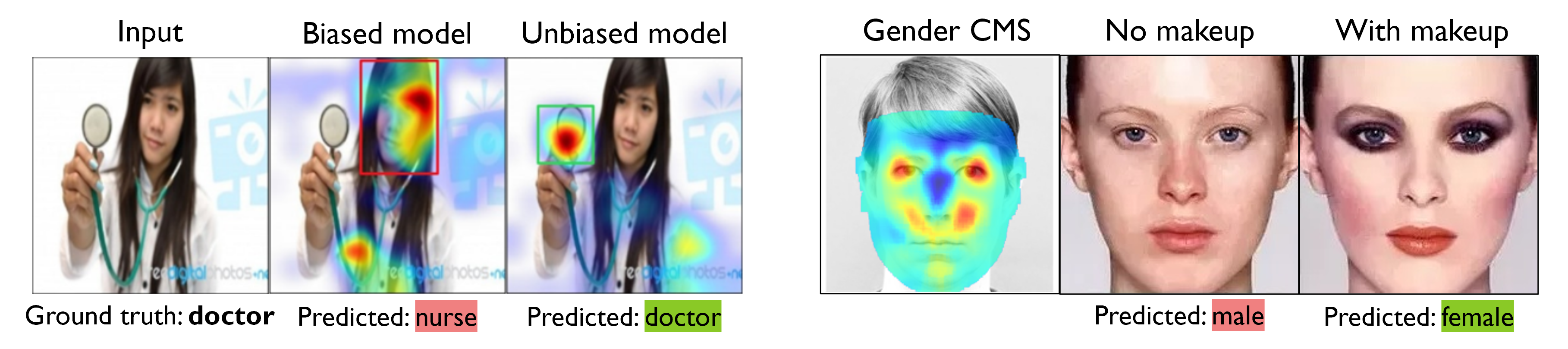}
	\vspace{-20pt}
	\caption{How are saliency maps useful for understanding incorrect predictions? Here we show examples in case of two biased predictions. \textit{(Left:)} GradCAM is used to show that a biased model focused on the face and appearance features to classify occupation, whereas an unbiased model used the tools in the input image (Image adapted from \cite{selvaraju2017gradcam}). \textit{(Right:)} A biased model classifies a woman’s face as male when she is not wearing makeup. The CMS (saliency) map shows that the model relies on the eyes for gender classification (Image adapted from \cite{john2021canonical}).}
	\label{fig:face_bias_saliency}
\end{figure}

We begin by discussing one of the most well-known post-hoc visualization methods: \textit{saliency maps}.
All input features do not have equal importance to a model's prediction. Saliency algorithms attribute the model's final decision to the input features. They are usually presented as a heat map of attribution over input features, called a saliency map or attribution map. The objective of saliency maps is to reveal the rationale behind model decisions. Observing  salient input features can also help uncover implicit biases in a model. If the prediction is wrong, saliency maps may help pinpoint the features that resulted in the error. Selvaraju et al. demonstrated using GradCAM \cite{selvaraju2017gradcam}  that a biased model may rely more on facial and appearance features while classifying occupation, whereas an unbiased model focuses on the tools (see Figure \ref{fig:face_bias_saliency}). Saliency maps also help us explore the nature of a task and the discriminative parts of an image concerning a class. For example, in face recognition, certain facial features of a person may resemble someone else. While recognizing emotions, the eyes may frown while the mouth is smiling, as shown in Figure \ref{fig:titlefig}. Generic image saliency algorithms do not always work on face images. For example, Figure \ref{fig:face_gradcam} shows that GradCAM is bad at highlighting features \emph{within} a face.

\red{Quantitatively, a saliency map has to be \emph{faithful} i.e. the heatmap should accurately reflect the ‘real’ attribution of features. Qualitatively, the saliency map has to be \emph{helpful} i.e. humans can derive actionable insights from them. }

\red{There are three considerations for saliency algorithms: \textit{helpfulness} \cite{doshi2017towards,doshi2018considerations} or utility for humans to make decisions, \textit{trustworthiness} \cite{silva2018towards} or reliability and \textit{fidelity} (discussed in detail in Section \ref{sec:saliencyevaluation}). Doshi-Velez and Kim \cite{doshi2017towards, doshi2018considerations} mention five factors related to the usefulness of explanations to humans, in essence measuring the simplicity of the explanations and how intuitive they are to interpret:
	\begin{itemize}
		\item \emph{Form of cognitive chunks.} What are the basic units of the explanation? (input pixels, collections of pixels, concepts?)
		\item \emph{Number of cognitive chunks.}  How many cognitive chunks does the explanation contain? \item \emph{Level of compositionality.}  Are the cognitive chunks organized in a structured way?
		\item \emph{ Monotonicity and other interactions between cognitive chunks.} Does it matter if the cognitive chunks are combined in linear or nonlinear ways? 
		\item \emph{Uncertainty and stochasticity.} How well do people understand uncertainty measures? \end{itemize}}

\red{ Silva et al. \cite{silva2018towards} introduce the ‘three Cs of interpretability’, which deal with the trustworthiness of an explanation, in addition to its simplicity: }
\begin{itemize}     
	\item \red{\emph{Completeness: } Users should be able to apply the explanation to cases where it is known and can validate it.  } 
	\item \red{\emph{Correctness}:  The explanation should be accurate. }   
	\item \red{\emph{Compactness: } The explanation should be succinct. This condition was related to rule-based explanations. Regarding saliency maps, this can mean that the level of detail and number of discontinuous chunks should be manageable. }
\end{itemize}

We broadly classify existing saliency algorithms into perturbation-based and backpropagation-based algorithms, of which gradient-based algorithms are a subclass. Figure \ref{fig:face_attention_survey} shows a comparison of saliency maps for faces. Below, we present an overview of some popular saliency algorithm classes.

\subsection{Perturbation-based Saliency Maps}
Perturbation-based methods find the saliency of the input features compared to a baseline sample by perturbing input features and observing the effect on output. These algorithms are architecture and implementation-agnostic but computationally expensive, as they pass the input sample through the model several times with different perturbations.

\begin{figure}
	\centering
	\includegraphics[width=\linewidth]{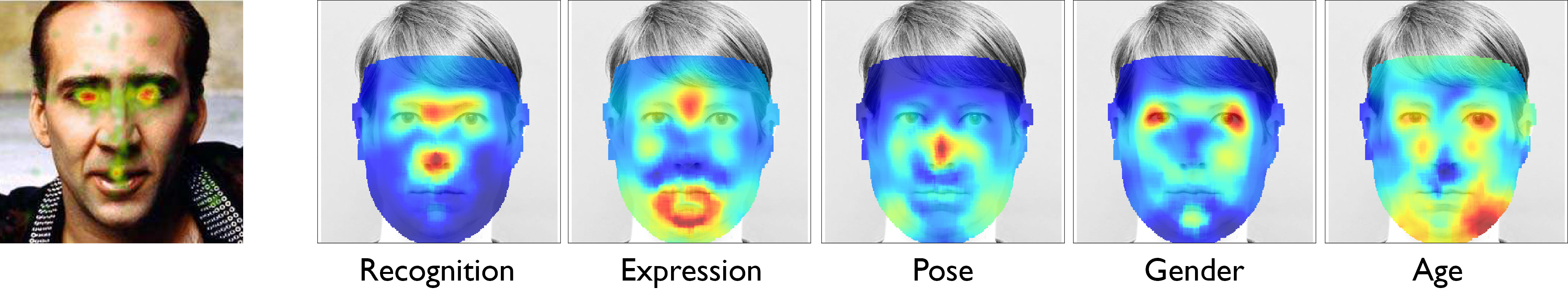}
	\vspace{-20pt}
	\caption{Left column shows the fixation heatmap when humans were asked to freely view a face image (\cite{Xu2015LearningGaze}). Other columns show aggregated saliency maps for different face tasks calculated using Canonical Model Saliency Maps \cite{john2021canonical}.}
	\label{fig:face_saliency}
\end{figure}

Occlusion maps \cite{zeiler2014visualizing} systematically slide a window over an input sample, flipping the pixels to the baseline image and observing the change in output class confidence. The saliency of a patch is given by:
\begin{equation}
	H_p = f(x) - f(x \odot (1-p))
\end{equation}
where $x$ is the input sample, $p$ is the patch and $f(\cdot)$ is the class confidence. Occlusion maps produce interpretable and intuitive maps whose coarseness can be controlled by the window size. Zhong and Deng \cite{Zhong2019Exploring} used occlusions at predefined locations to calculate the importance of coarse facial features for face recognition. They also highlighted the difference between similar faces by occluding the same features of two images at once and calculating the feature distance \cite{zhong2018difference}. John et al. systematically occluded the input face image and projected the resulting saliency map onto a neutral frontal face \cite{john2021canonical}. This ‘canonicalization’ procedure allowed them to collate multiple saliency maps to obtain model-level saliency maps (See Figure \ref{fig:face_saliency}).  

Shapley values \cite{shapley201617} provide a theoretical framework for perturbation-based methods which obeys the axioms of completeness, symmetry (two features must be attributed equally if they have an equal effect on the output) and sensitivity. Shapley values consider the input features to be players in a coalition game where the outcome is the payout. The algorithm fairly distributes the payout between each player based on their contribution. Given a model $f$ and $n$ features, the payout for a feature $i$ is computed as:
\begin{equation}
	\phi_i(f) = \mathlarger{\sum}_{S \subseteq \{1 ... n\}/i} \frac{|S|!(n-|S|-1)!}{n!} \big(f(S \cup \{i\}) - f(S)\big)
\end{equation}
where $S$ is a subset of the features and $x$ is the vector of feature values of the input instance. While Shapley values satisfy many theoretical constraints, they are not practical to calculate for input with many features like image data. Variations such as SHAP and Kernel SHAP \cite{lundberg2017unified} make the computation more practical. 

Local Interpretable Model-Agnostic Explanation (LIME) \cite{ribeiro2016should} uses local surrogate models to explain individual predictions. LIME generates a dataset by perturbing the input sample's features and recording the model's corresponding predictions, given a model and an input sample. LIME then trains an interpretable model on this generated dataset. This learned model is a good approximation of the model locally but not globally (local fidelity).

\subsection{Backpropagation-based Saliency Maps}

Backpropagation-based algorithms start with a trained model and a desired output. A high gradient at the desired output is then backpropagated to the model elements, starting with the layers closest to the output, all the way down to the input. This procedure finally assigns a score to each input feature based on their contribution to the output. They are easy to calculate as they require only a single backward pass. However, they tend not to be implementation-agnostic by their very nature. 

Layer-wise relevance propagation (LRP) \cite{bach2015LRP} uses the structure of layered neural networks to assign prediction score to elements of a layer, ensuring that the relevance flowing into a unit should equal the relevance flowing out to a lower layer. If we observe column (h) of Figure \ref{fig:face_attention_survey}, we notice that the output of LRP on faces visually resembles guided backpropagation. LRP tends to highlight image edges, and in most cases, it highlights all facial features equally. The high number of separate salient areas makes it difficult to interpret. 

Excitation Backprop or EBP \cite{zhang2018excitationbackprop} uses a probabilistic winner-takes-all formulation. Let the relevance of a prediction be specified by the prior distribution $P(A_0)$ over the output neurons. Let $P(A_t|A_{t-1})$ be the probability of selecting neuron $A_t$ in layer $t$ as the winning neuron given $A_{t-1}$ is selected in the layer before. We calculate the marginal winning probability as:
\begin{equation}
	P(a_i) = \sum_{a_j \in P_i} P(a_i|a_j)P(a_j)
\end{equation}
which gives the relevance of each neuron. Here, $a_i$ denotes a specific neuron, and $A_t$ refers to a variable over the neurons). This work specifies the conditional winning probability based on the weights between each neuron and the activation. Contrastive EBP (c-EBP) is a variation that calculates the contrastive saliency between pairs of classes. Column (j) of Figure \ref{fig:face_attention_survey} shows the results of applying unmodified EBP on face images. We observe that it highlights almost the entire face. Thus, unmodified EBP does not give actionable insights on face images. Researchers have proposed modified versions of EBP designed to work on faces. Castanon and Byrne \cite{castanon2018Visualizing} used a variant of Excitation Backprop' (EBP)  and 'contrastive Excitation Backprop (cEBP) \cite{zhang2018excitationbackprop} called 'truncated cEBP' to compute the network saliency on faces, where the network attention signal is propagated from the output neurons back to the input pixels. Williford et al. \cite{williford2020explainable} focused on explaining the matches returned by a facial matcher to understand why a probe was matched with one identity over another. The unit for explanation is a triplet of (probe, matching face and non-matching face). They adapted EBP, cEBP and tcEBP by using triplet loss instead of cross-entropy loss.  

DeepLIFT \cite{shrikumar2017deeplift} breaks down the difference between a neuron's output and reference output to all the upstream neurons connected to it such that the sum of contributions equals the difference between output and reference output. Column (i) of Figure \ref{fig:face_attention_survey} shows DeepLIFT applied to faces. Compared to LRP, DeepLIFT has fewer salient areas highlighted and thus is easier to interpret. However, the results do not seem to agree with other saliency map methods. 

\subsection{Gradient-based Saliency Maps}

The gradient of the output with respect to an input pixel represents how much difference a tiny change in the pixel would make to the output. Thus it may be used to highlight salient pixels \cite{simonyan2013deep}. Gradient-based saliency maps are noisy and discontinuous due to the 'shattered gradient effect' \cite{balduzzi2017shattered}. For deep rectifier networks, the discontinuities in the gradient increase exponentially with number of layers. Also, the gradient values change much more rapidly than the corresponding change in input \cite{samek2021survey}, adding more noise to the heatmap. Moreover, these methods are not implementation-independent. 

Smilkov et al. proposed an improvement called 'SmoothGrad' \cite{smilkov2017smoothgrad}, which attempted to reduce the noise by averaging over several saliency maps of the same input image, each produced after adding some random noise to the input image. Column (b) of Figure \ref{fig:face_attention_survey} shows that the results are highly discontinuous and challenging to interpret even after smoothing. Gradient-based methods break the axiom of 'Sensitivity' as the prediction function may flatten at the input and have zero gradients despite the function value at the input being different from the baseline. Sundararajan et al. proposed 'Integrated Gradients', which preserves sensitivity. They start with a baseline image and calculate the gradient at equally spaced intervals on the straight-line path between the baseline image and the input image. They then integrate these gradients to get the final saliency heatmap. The integrated gradient along the $i^th$ dimension is given by:
\begin{equation}
	H_i = (x_i - x^0_i)  \int_{\alpha=0}^1 \frac{\partial f(x^0 + \alpha(x-x^0))}{\partial x_i}
\end{equation}
where $x$ is the input sample and $x^0$ is the baseline image. In practice,  integrated gradients look similar to SmoothGrad with many disparate areas (Column c). Although there are theoretical guarantees, our experiments showed it to not be as useful for the face domain, where one object dominates the entire image. 

Class Activation Mapping (CAM) is a category of gradient-based saliency algorithms which focuses on localizing the objects of importance in the input image. Zhou et al. \cite{zhou2016CAM} proposed highlighting salient objects without discontinuities for architectures where the feature maps directly preceded the softmax layers. Selvaraju et al. \cite{selvaraju2017gradcam} extended the algorithm to cover a wide range of architectures in their popular method, '\textit{Grad-CAM}'. Given a class $c$ and an output neuron $y^c$, it sums the weighted feature maps $A_k$ of a layer of interest: 
\begin{equation}
	H^c = ReLU \Big(\sum_k \alpha_k^c A^k \Big)
\end{equation}
where each feature map is weighted by the gradient of the class output with respect to the activation maps as follows:
\begin{equation}
	\alpha_k^c = \frac{1}{N} \sum_i \sum_j \frac{\partial y^c}{\partial A^k_{ij}}
\end{equation}
Finally, the map is upsampled using bilinear interpolation to match the input image's size. Many extensions of Grad-CAM exist, such as GradCAM++ \cite{chattopadhyay2018gradcamplus} and ScoreCAM \cite{wang2020scorecam}. Columns (d) and (e) of Figure \ref{fig:face_attention_survey} show the results Grad-CAM and ScoreCAM on faces. Although they successfully localize the object of interest, they do not provide finer details, rendering them not as useful for face images (See also Figure \ref{fig:titlefig}). 

\red{Luo et al. \cite{Luo2016ERF} introduced a related concept called `effective receptive field' (ERF), defined as the region of the input image containing any pixel with a non-negligible impact on the central output unit. The impact of a single input pixel $x_{i,j}$ on the central output unit $y_{0,0} $is measured by the partial derivative $\frac{\partial y_{0,0}}{\partial x_{i,j}}$. They define the ERF as the expectation over the input distribution of the impact of each input pixel. They discovered that the effective receptive field occupies only a fraction of the entire theoretical receptive field and that, in many cases, ERF follows a Gaussian distribution.}

\subsection{Insights on Face Saliency}
\label{sec:saliencyinsights}

\red{\textbf{Generic vs. face-specific saliency: } Faces exhibit a distinctive shared structure, with features such as a central nose, mouth below the nose, and eyebrows above the eyes. Successful face explainability algorithms acknowledge this inherent structure while accommodating pose variations. Yin et al. \cite{yin2019Towards} proposed an intrinsically interpretable face recognition model, trained with image pairs wherein one had a precisely occluded facial region using face alignment. Williford et al. \cite{williford2020explainable} introduced an evaluation metric for face explainability,which uses a curated dataset with  select facial regions manually altered. Likewise, Zhong and Deng \cite{zhong2018difference} utilized cosine similarity to compare face pairs with the same regions occluded using face alignment. Respecting the facial structure is critical in adapting generic saliency algorithms to face-specific ones. Williford et al. \cite{williford2020explainable} achieved this by transforming the generic explainability algorithm RISE \cite{petsiuk2018rise} into the face-specific DISE by employing random occlusions focused on important facial regions. John et al. \cite{john2021canonical} converted generic occlusion maps to the face-specific CMS by occluding a face with reference to a canonical face and mapping the resulting confidence drop back to the canonical face.}

\textbf{Patterns of saliency on face images:} Face models focus on different facial features depending on their task. Figure \ref{fig:face_saliency} shows the focus patterns for different face tasks in comparison to the focus pattern of the human gaze. John et al. \cite{john2021canonical} discussed the implications of these focus patterns and insights into the tasks. Recognition focuses on the eye-nose triangle, whereas gender uses the corners of the eyes (see also Figure \ref{fig:face_bias_saliency}). Xu et al. \cite{Xu2015LearningGaze} recorded the gaze of humans when freely viewing faces. We notice that this resembles the focus patterns of recognition models. \cite{sadr2003eyebrows} showed that eyebrows are most instrumental in the human recognition of faces. 

\vspace{5pt}
\noindent \textbf{Explainability for assisting humans:} While most existing explainability works focus on explaining how models work or why they make certain decisions, we now discuss explainability methods that aid humans in verifying models' decisions or help them make better decisions. Zee et al. \cite{zee2019enhancing} used existing explainability methods to enhance human face recognition performance, specifically the ability to distinguish between two similar-looking celebrity faces. They trained CNNs for the recognition and verification of these two identities. Using explainability visualizations, they found that the changes between the two identities are primarily on the forehead, cheeks and lower face and not on the eyes or the mouth. On instructing novice participants to focus on the forehead and cheekbones, the participants had higher accuracy in distinguishing between the identities. Zhong and Deng \cite{zhong2018difference} created a visualization method to assist human evaluators in identifying people who try to invade the biometric system using similar-looking faces. Given a pair of aligned faces, they systematically occluded portions of the face pair and mapped the resulting drop in cosine distance to a heatmap. They then normalized this heatmap with a learned threshold to highlight differences in negative pairs but not in positive pairs. Their visualization enabled human evaluators to increase their accuracy from 76\% to 83\%.

\vspace{5pt}
\noindent \textbf{Level of detail in saliency maps:} Figure \ref{fig:face_attention_survey} compares different saliency map methods on the VGG-Face model for the task of face recognition. The figure shows that different saliency map algorithms highlight different levels of detail. Integrated Gradients, Guided Backprop, and DeepLIFT show fine detail and resemble edge detectors. Class activation mapping-based approaches are coarse (we used the layers typically used in well-known efforts for these visualizations). Class-level saliency maps are not very informative for face images, each of which generally only contains one object (face) instance. Algorithms that show mid-level detail, like EBP and CSM, seem to be in the Goldilocks zone since the features we are interested in (parts of the face) are of that size. So, it is crucial to pay attention to the level of detail while designing saliency algorithms for faces. 

\subsection{Evaluation of Saliency Maps}
\label{sec:saliencyevaluation}
To measure the fidelity of a saliency map, we often assesses the attribution of each input feature compared to a ‘baseline input’, which is a neutral input that could also have zeros for all features. The baseline input for an object recognition task may be a completely black image, whereas it may be a zero vector for a word embedding task.   
Sundarajan et al. \cite{sundararajan2017axiomatic} provided three axioms to evaluate the fidelity of an explanation: (i)	\emph{Sensitivity}: the attribution of a feature should reflect a network’s change in output on the inclusion or exclusion of the feature. (ii)	\emph{Implementation Invariance}: the attribution of two functionally invariant models should be the same for the same inputs. (iii)	\emph{Completeness}: the sum of all attributions should be equal to the difference in prediction between the input and baseline.
Below, we present some quantitative measures used to evaluate the fidelity and trustworthiness of saliency algorithms.

\vspace{5pt}
\noindent \textbf{Mask and evaluate:} Chattopadhyay et al. \cite{chattopadhyay2018gradcamplus} introduced a protocol in which unimportant areas of an input image, as identified by a saliency map, are masked, and the resulting drop in prediction confidence is measured. This method does not translate well to faces as, unlike object recognition, face images have a single object at the center of the image, and models trained on face images focus on different parts of the faces. For face images, John et al. \cite{john2021canonical} instead mask important image parts and observe the confidence drop to address this issue in face images. They also normalize the sum of heatmap pixels so saliency algorithms cannot 'cheat' by covering a large area.

\vspace{5pt}
\noindent \textbf{Hiding game: } This is a variation of 'Mask and evaluate' where the pixels are gradually flipped in order of their saliency \cite{castanon2018Visualizing}. This game is also called `pixel flipping' \cite{samek2016evaluating} or `insertion and deletion metrics' \cite{wang2020scorecam}. The insertion metric starts with a baseline image and gradually adds the most salient pixels. The speed of increase in confidence measures the quality of the saliency map. The deletion metric starts with the input image and gradually replaces the most salient pixels with baseline image pixels. We then measure how quickly the confidence falls as compared to other saliency methods or random flipping of pixels.

\vspace{5pt}
\noindent \textbf{Pointing game:} This is a supervised test that measures how well a saliency method localizes relevant regions of an image. This procedure extracts the maximum point of the saliency map and checks whether it falls within the ground truth bounding box of the object \cite{zhang2018excitationbackprop}. Wang et al. \cite{wang2020scorecam} extended this measure by adding up all heatmap pixels that fall inside the object's bounding box. A major drawback of this method is that we cannot know a priori if the model's ground truth reasoning does not match human expectations. Also, it works on simple tasks where we expect salient points to fall inside the object strictly \cite{kim2021sanity}. 

\vspace{5pt}
\noindent \textbf{Inpainting game for faces \cite{williford2020explainable}: } Williford et al. constructed a curated database by inpainting some predefined features (such as nose, left eye, left eyebrow, etc.) with features from another identity. They create a customized dataset for each network by choosing images such that the network can distinguish the probe image and inpainted image. The inpainting game is played as follows: The saliency algorithm is presented with a triplet of the probe, mate and inpainted non-mate, and is tasked with estimating a discriminative saliency map that estimates the likelihood that a pixel belongs to a region that is discriminative for the mate. The pixels classified as being salient by sweeping the saliency threshold are replaced with the pixels from the "inpainted probe". These "blended probes" can then be classified as original identity or inpainted non-mate identity by the network being tested. This measure is suitable for face models. High-performing deep learning models will correctly assign more saliency for the inpainted regions that will change the identity of the blended probes without increasing the false alarm rate of the pixel salience classification. 

\vspace{5pt}
\noindent \textbf{Randomised sanity checks: } This measure evaluates the trustworthiness of the saliency visualization. Adebayo et al. \cite{adebayo2020sanity} showed that many saliency maps resembled edge detector outputs. They proposed two sanity checks to ensure the correctness of the saliency algorithms: \emph{Model parameter randomization test} and \emph{Data randomization test}. 
\emph{Model parameter randomization test} checks if the saliency map depends solely on the input image. To this end, it calls for comparing the saliency map on a trained model with a saliency map generated on a randomly initialized model. If the algorithm depends on the learned parameters of the model, there should be a substantial difference between the two maps. If we randomize the weights from the top layer to the bottom layer, we expect the saliency map to be progressively more random.
\emph{Data randomization test} compares the saliency map applied to a model trained on a labeled dataset with one trained on the same dataset with the labels randomly assigned. If the saliency method depends on the labeling of data, we should expect the saliency maps to be drastically different.

We observe from the above discussion that just like we need face-specific saliency visualization methods to capture the subtleties of a face processing task, one needs to also carefully choose evaluation metrics that are relevant to face images. 
\section{Feature Visualization in Face Models}

\label{sec:featurevisualisation}

Human vision is adept at breaking scenes into semantic components like objects and object parts. Similarly, we may break down faces as composed of facial features like eyes, nose and mouth. A natural question when visualizing DNN-based face processing models is whether the hierarchical patterns learned by filters of convolutional layers match human intuition. What is the difference between the patterns learned on natural images versus those learned on faces? Do these visualizations differ between face tasks like recognition and pose? Feature visualization methods attempt to understand DNN model features by projecting them onto the input domain. Apart from understanding the hierarchy of learned filters, such methods also help us answer questions such as: Are there redundant features or layers? Do the DNN learn biases or incorrect associations from the dataset? It was shown in \cite{googlenetvis2016} how visualizations of the ‘\textit{saxophone}’ class also included the musician holding the instrument. Feature visualization methods can also help us detect such entanglement or correlations between context and content. We begin our discussion with a detailed background of face visualization methods, and subsequently show results with well-known face models on face images.

\begin{figure}
	\centering
	\includegraphics[width=\linewidth]{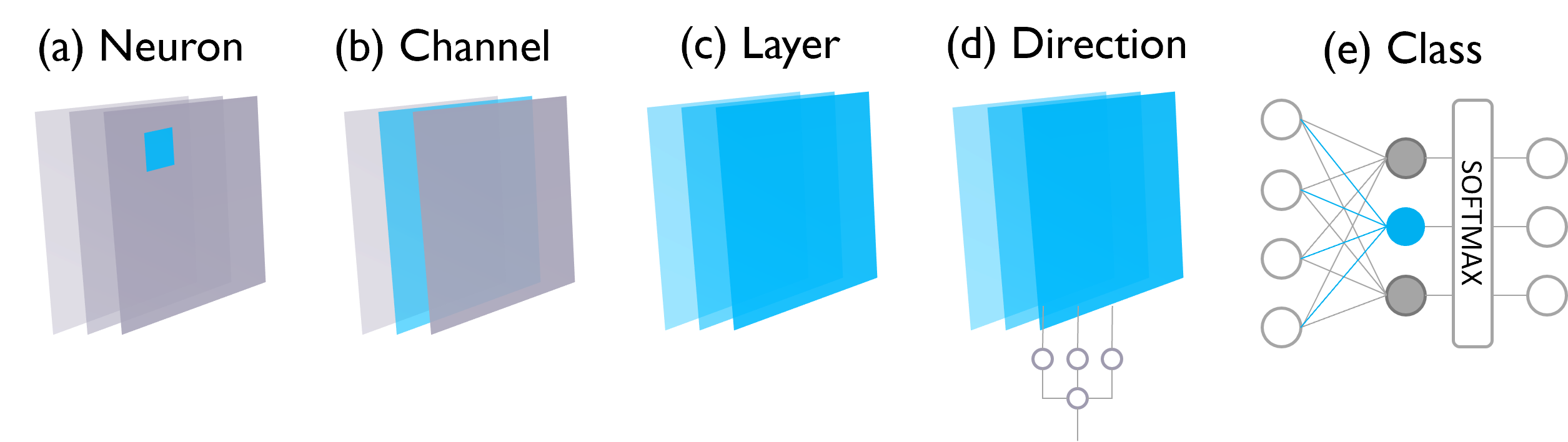}
	\vspace{-18pt}
	\caption{We visualize features at different granularities: neurons at an individual position, an entire channel, linear combinations of channels, or an entire layer. We can also visualize classes (inspired by \cite{olah2017feature}).}
	\label{fig:visualization_units}
\end{figure}

\begin{figure*}
	\centering
	\includegraphics[width=\linewidth]{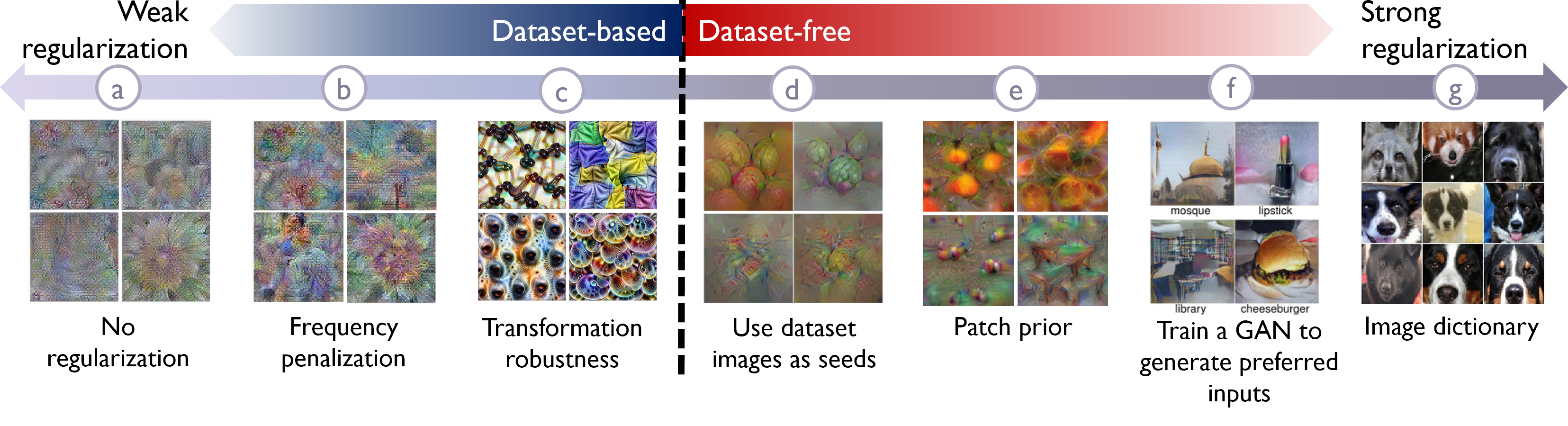}
	\vspace{-20pt}
	\caption{Spectrum of regularizations for feature visualization. On the left are methods that perform visualizations without access to data. The methods on the right take help from a dataset in some form. (\textit{Image sources:} (a), (b), (c) and (g) are from \cite{olah2017feature}; (d) \cite{nguyen2016multifaceted} (e) \cite{wei2015intraclass} (f) \cite{nguyen2016synthesizing}).}
	\label{fig:visualization_regularization}
\end{figure*}

\subsection{Background: Feature Visualization Methods}

Early efforts in this direction visualized higher-layer features as linear combinations of lower-layer filters \cite{lee2009convolutional}. However, such an approach does not consider the non-linearities in the network. A more flexible formulation considered feature visualization as a ‘search’ for input images that cause a neural unit to give a high response. Erhan et al. \cite{erhan2009visualizing} introduced \emph{activation maximization}, which optimizes an input image using gradient ascent so that the activation or response of a neural unit is maximized. This has since been a popular method used for feature visualization. Newer works study the effects of various regularization terms to mitigate some shortcomings of the original formulation. Below is a general expression for feature visualization:
\begin{equation}
	x^* = \argmax_{x \in X} \phi(x) - \lambda R
	\label{eq:activation_maximization}
\end{equation}
where $X$ is the search-space of the input, $\phi(x)$ is the activation of the neural unit under study and $R$ is a term that regularizes for the quality or diversity of the visualization. In practice, the search-space $X$ is mostly $\mathbb{R}^{W \times L \times C}$ or crops of dataset images $\mathcal{T}$ \cite{zeiler2014visualizing}.

As shown in Figure \ref{fig:visualization_units}, the neural unit under study, $\phi$, can refer to a channel at a specific position on the input image or an entire channel \cite{simonyan2013deep} . We can also search for inputs that maximally activate any neuron in a layer as in Figure \ref{fig:visualization_units}(c) \cite{mordvintsev2015deepdream, gatys2016styletransfer}. This shows the patterns that each convolutional filter 'looks for' in an input image. The algorithm can pick any of the layers’ channels to maximize its activation at every position in the image. This procedure results in dream-like visualizations where features of interest are enhanced. Channel visualizations are motivated by the assumption that individual channels form a distinguished basis which is particularly useful for extracting semantic information. (See Section \ref{sec:concepts}). Jointly optimizing multiple channels \cite{olah2017feature} gives us information about how the neurons interact. Visualizing one of the output neurons (before softmax) is often called ‘class visualization’ (Figure \ref{fig:visualization_units}(e)), which finds a representative image for an entire class.

Unconstrained activation maximization can lead to adversarial images \cite{szegedy2013intriguing}  or visualizations with noise and unwanted high-frequency signals \cite{odena2016checkerboard}. Hence, recent feature visualization methods focus on regularizers that add a ‘natural image prior’ to the objective function so that the visualization falls in the dataset space $\mathcal{D}$. As shown in Figure \ref{fig:visualization_regularization}, regularization functions exist on a spectrum from strong to weak. Dataset-free regularizations attempt to mimic statistics of natural images by penalizing high frequencies \cite{simonyan2013deep, olah2017feature, yosinski2015understanding, googlenetvis2016,tyka2016bilateral } or adding transformation robustness \cite{ olah2017feature, googlenetvis2016}. Dataset-based regularizations tend to have a better quality of visualizations at the cost of missing the space of $\mathcal{D} - \mathcal{T}$. The easiest method picks image parts from the training dataset with the highest filter response \cite{zeiler2014visualizing, wei2015intraclass, olah2017feature}. Nguyen et al. \cite{nguyen2016synthesizing} trained a Generative Adversarial Network (GAN) on the input dataset and searched the GAN latent space for an appropriate image. Other works seed the visualization with dataset images \cite{nguyen2016multifaceted} or patches of dataset images \cite{wei2015intraclass}. Special diversity terms in the regularization ensure that activation maximization unearths multiple facets of a filter \cite{olah2017feature, nguyen2016multifaceted}.

\subsection{Features of Facial Models}

Before we present our results on face images, we hypothesize three desired properties of a good feature visualization, especially for a domain like face images: The visualization should be \emph{interpretable in-domain}. It should not have \emph{ undue influence from a limited dataset} and it should \emph{show all facets of the unit under study}. Let $\mathcal{D}$ be the domain of images ‘expected’ as input by a deep network or parts of such images. For object detection, $\mathcal{D}$ may be natural images; for face recognition,  $\mathcal{D}$ may be images of faces and parts of the face, and so on. Let $\mathcal{T} \subseteq \mathcal{D}$ be the training set of the network. $\mathbb{R}^{W\times L} - \mathcal{D}$ contains out-of-domain and adversarial images. Ideally, we prefer our visualization to be in $\mathcal{D}$ without being restricted to $\mathcal{T}$. We also want our visualization to capture the diversity of patterns unit in the network ‘looks for’.  

\begin{figure}
	\centering
	\includegraphics[width=\linewidth]{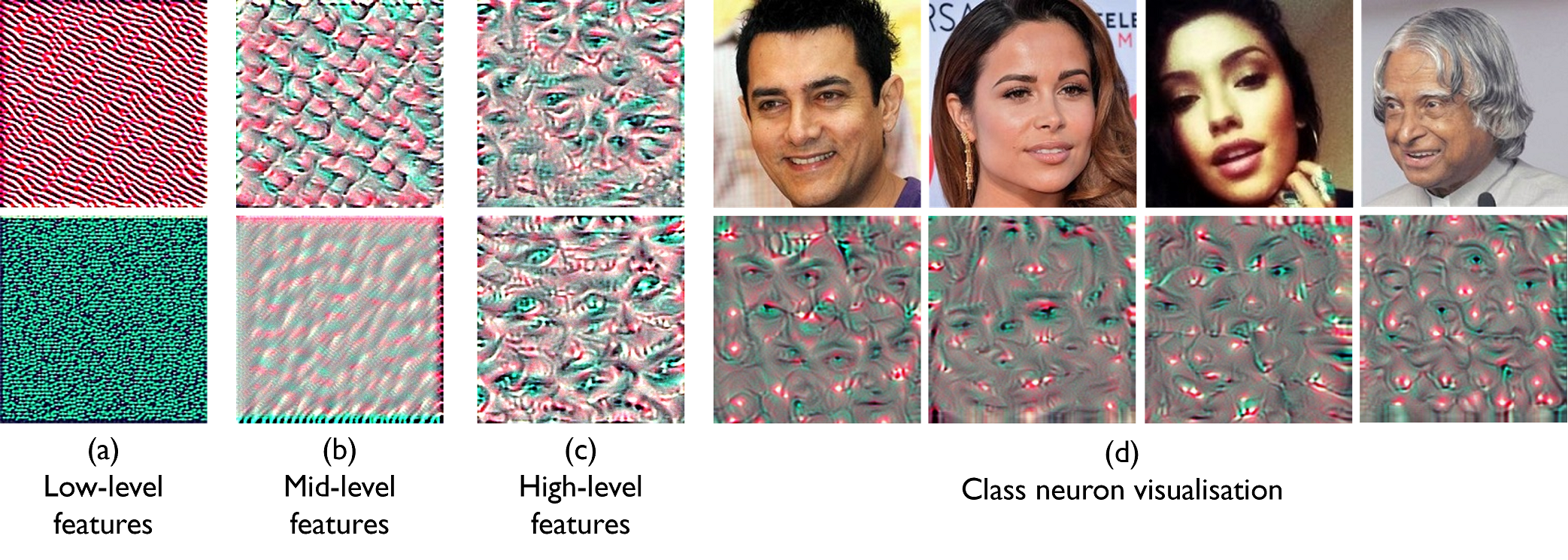}
	\vspace{-17pt}
	\caption{Visualizations of internal representations of the VGG-Face network \cite{parkhi2015vggface}. (a) Early conv layers (1 to 6) show simple patterns. (b) Patterns in the middle layers (7 to 9) show more complex patterns. (c) Later convolutional layers (10 to 13) show facial features and parts of the face. (d) Class neurons capture facial identities. These visualizations were created using activation maximization algorithm with L-2 regularization.}
	\label{fig:facefeatures}
\end{figure}

\begin{figure}
	\centering
	\includegraphics[width=\linewidth]{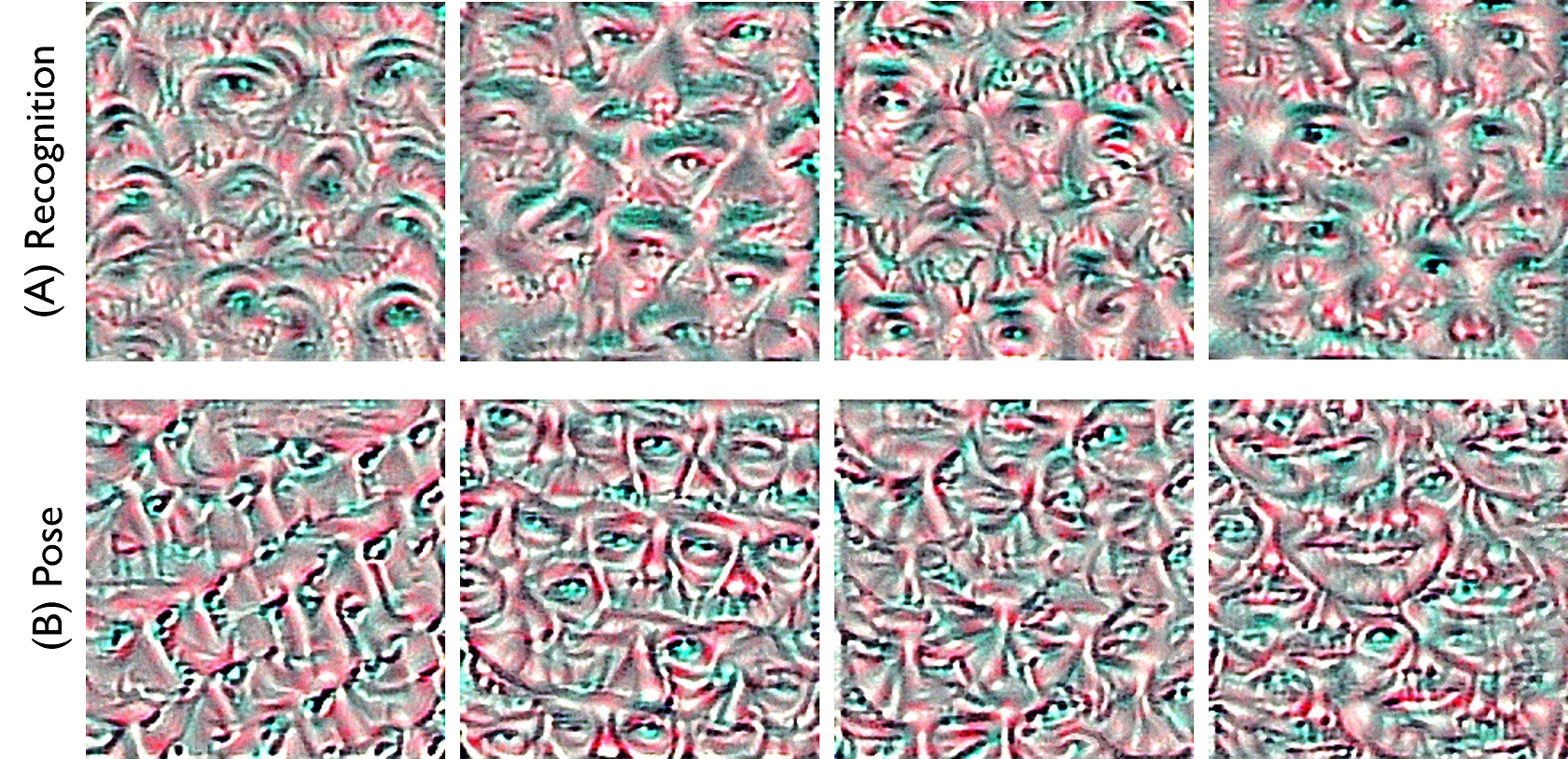}
	\caption{Sample patterns that activate filters in the last convolution layer of the VGG-Face recognition network \cite{parkhi2015vggface} (top row) and a head pose VGG-16 network (bottom row) . The recognition network's filters show different shapes of eyes and nose, each probably representing a different identity. The pose network filters show variation in the pose of the features instead of their shapes.}
	\label{fig:facefeaturenetworks}
\end{figure}

Figure \ref{fig:facefeatures} shows that deep face models learn hierarchical features from face images. The initial layers learn simple patterns similar to object classification networks, while higher layers learn complex patterns composed of these simple patterns. The middle levels learn facial features such as eyes and nose. Neurons in the fully connected layers learn parts of the face, while the last level neurons learn representations of the entire face. Mirroring this observation, Zhong and Deng \cite{Zhong2019Exploring} created a dictionary of feature visualizations using Deconvnet \cite{zeiler2014visualizing} showcasing the hierarchy in facial features. 

Figure \ref{fig:facefeaturenetworks}, on the other hand, shows that various face networks learn different features based on their tasks. Various filters in the higher layers of a recognition network learn patterns corresponding to different shapes of eyes and nose, as these features help to tell identities apart. In the higher layers of a head pose network, there is not much variation in the shapes of features. However, they contain parts of the face in different poses. We observed across our studies that \textit{part geometry} and \textit{task objective} are two complementary elements that face model visualizations highlight in general.

\section{Feature Inversion in Face Models}
\label{sec:featureinversion}

\begin{figure}
	\centering
	\includegraphics[width=\linewidth]{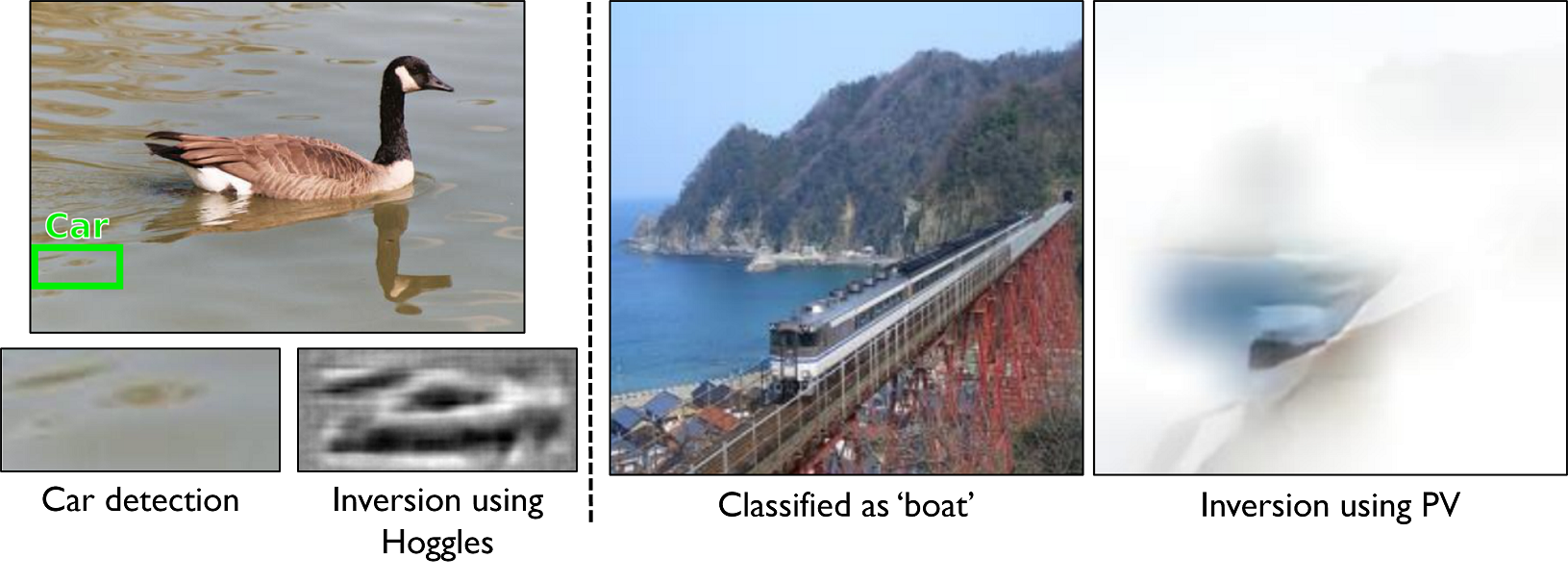}
	\vspace{-14pt}
	\caption{Examples of deep network `mistakes' that are explained using feature inversion. \textit{Left:} HOGgles \cite{vondrick2013hoggles} is used to explain a wrong `car' detection by HOG features. \textit{Right:} Perception Visualization (PV) \cite{giulivi2021perception} is used to explain a wrong `boat' classification using deep features.}
	\label{fig:inversion_mistake}
\end{figure}

In the previous section, we observed how feature visualization allows us to see the 'bases' of a convolutional neural network (CNN) typically used in deep face models. Feature inversion is a closely related approach in which we project the deep features corresponding to an input image back to the input domain. This method allows us to look at the input image through the same 'lens' as the deep network. As we feed an image through a deep face network, each layer abstracts the output of the preceding layers until we reach the final classification layers by retrieving features discriminative towards crucial characteristics and invariant to undesirable factors like illumination and pose in each layer. Feature visualization reveals which information is kept or discarded by each layer, revealing the aspects of the images that are discriminative for various tasks. 
On the other hand, feature inversion helps us identify why a network misclassifies or misinterprets a particular image (See Figure \ref{fig:inversion_mistake}).

Another use case for feature inversion is to ensure the privacy and security of users. Intelligent systems may store features or descriptors extracted from user images instead of saving the entire images to mitigate privacy concerns. Feature inversion helps detect the information stored in these descriptors and whether we can use them to reverse-engineer user-identifying systems such as faces or locations \cite{dangwal2021reverse}. Such visualizations can be different from the algorithms used for feature visualization. Feature inversion has also seen creative efforts such as Deep Dream \cite{mordvintsev2015deepdream, mordvintsev2015inceptionism} and Style Transfer \cite{gatys2016styletransfer}, which have been proposed in the past to generate digital art from images. 

\subsection{Background of Feature Inversion Methods}
We may formulate feature inversion as follows: Let $x^0$ be an input image and $\phi(.)$ be a function that extracts features from the desired layer of a deep network. Let $\phi^0 = \phi(x^0)$ be the given features to be inverted. We need to find the 'pre-image' $x^*$ such that its features $\phi(x^*)$ is close to the given features $\phi^0$. Similar to Equation \ref{eq:activation_maximization}, we may formulate feature inversion as an optimization function that finds an image that minimizes the distance between its features and the given feature vector \cite{mahendran2015inverting}:
\begin{equation}
	x^* = \argmin_{x \in \mathbb{R}^{W \times L \times C}} L(\phi(x), \phi^0) + \lambda R
	\label{eq:feature_inversion}
\end{equation}
where $L$ is a loss function that determines the distance between features of the input image and the visualized image, usually $\norm{\phi(x) - \phi^0}_2$, $ R$ is a regularization function to ensure that the visualization meets the preferred characteristics for explainability.

Before discussing the results of feature inversion methods on deep face models, we discuss some desired characteristics of the inverted image $x^*$:

\begin{enumerate}
	\descitem{The closeness of $\bm{\phi(x^*)}$ to $\bm{\phi^0}$:} These features should be very similar to each other (or equal, in a ideal case) by definition of feature inversion.
	\descitem{The closeness of $\bm{x^*}$ to $\bm{x^0}$:} While this is not a required criterion for explainability, and we do not include in the objective function of Equation \ref{eq:feature_inversion}, we use feature inversion as a means to \emph{discern} if an inverted image is close to the original input image. As the abstraction of features at each layer causes the mapping between representations and input images to be many-to-one, the $x^*$ may not be identical to $x^0$ by default. 
	\descitem{$\bm{x^*}$ should be in the input image domain $\bm{\mathcal{D}}$}: If we directly optimize Equation \ref{eq:feature_inversion},  $x^*$ will be filled with high-frequency noise, similar to activation maximization. Existing methods hence use regularization functions that attempt to ensure that the generated image is in the input image domain $\mathcal{D}$. 
	\descitem{Using dataset images as input domain prior}: If the regularization term $R$ uses dataset images in some form, we must ensure that the dataset used is identical to the training dataset of the deep network.
\end{enumerate}

\begin{figure}
	\centering\includegraphics[width=\linewidth]{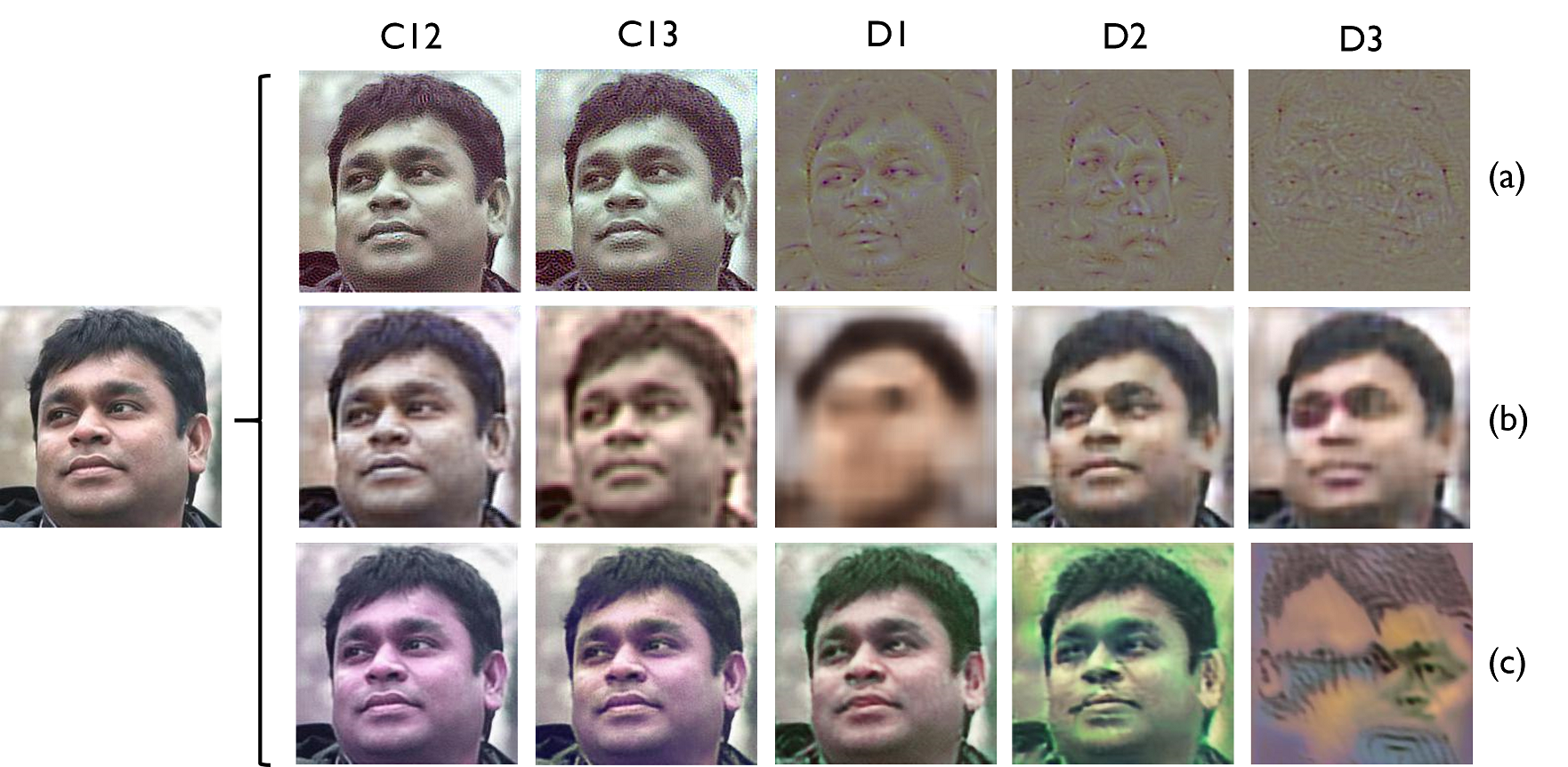}
	
	\caption{A face image from the VGG-Face dataset \cite{parkhi2015vggface} inverted using three different methods from the features of the last 5 layers of a VGG-Face network. C$x$ is the $x^{th}$ convolutional layer and D$x$ is the $x^{th}$ fully connected layer. The methods used: (a) Frequency penalization \cite{mahendran2015inverting}; (b) Inversion using CNN \cite{dosovitskiy2016inverting}; (c) Deep image prior \cite{ulyanov2018deep}.}
	\label{fig:face_inversion_methods}
\end{figure}
Like feature visualization, we get noisy and adversarial images if we directly optimize Equation \ref{eq:feature_inversion}. Previous research has proposed several regularization functions to bring the visualizations into the natural image domain. The regularizations are similar to those used for activation maximization, and both dataset-free and dataset-based approaches exist. Mahendran and Vedaldi \cite{mahendran2015inverting} directly penalized high frequencies by using $\alpha$-norm and TV norm. The top row of Figure \ref{fig:face_inversion_methods} shows that these regularizations are weak and cannot restore color information. Singh and Namboodiri \cite{singh2015laplacian} introduced Laplacian pyramids for regularization, with the intuition that a coarse-to-fine inversion scheme effectively recovers recognizable inversions of features from the different layers of a deep network. `Deep Image Prior' \cite{ulyanov2018deep} uses the structure of a convolutional generator network as a natural image prior. The bottom row of Figure \ref{fig:face_inversion_methods} shows that deep image prior produces high-quality reconstructions without using any information from the input dataset. Deconvnet \cite{zeiler2014visualizing} and guided backpropagation \cite{springenberg2015striving} exist at the intersection of feature visualization and inversion. They invert a single activation map instead of the full feature map of the layer. The simplest dataset-based method is to directly train a CNN to invert the features of each layer by giving feature-image example pairs from a dataset, as demonstrated by Dosovitskiy and Brox \cite{dosovitskiy2016inverting}. As we see in Figure \ref{fig:face_inversion_methods}, the resulting pre-images are blurry and heavily influenced by the dataset used to train the inverting networks. Instead of direct training from a dataset, other works \cite{weinzaepfel2011reconstructing, vondrick2013hoggles,wei2015intraclass} stitch together representative patches as a natural image prior. Dosovitskiy and Brox \cite{dosovitskiy2016generating} trained a GAN to invert features.

\subsection{Feature Inversion of Facial Models}
\begin{figure}
	\centering
	\includegraphics[width=\linewidth]{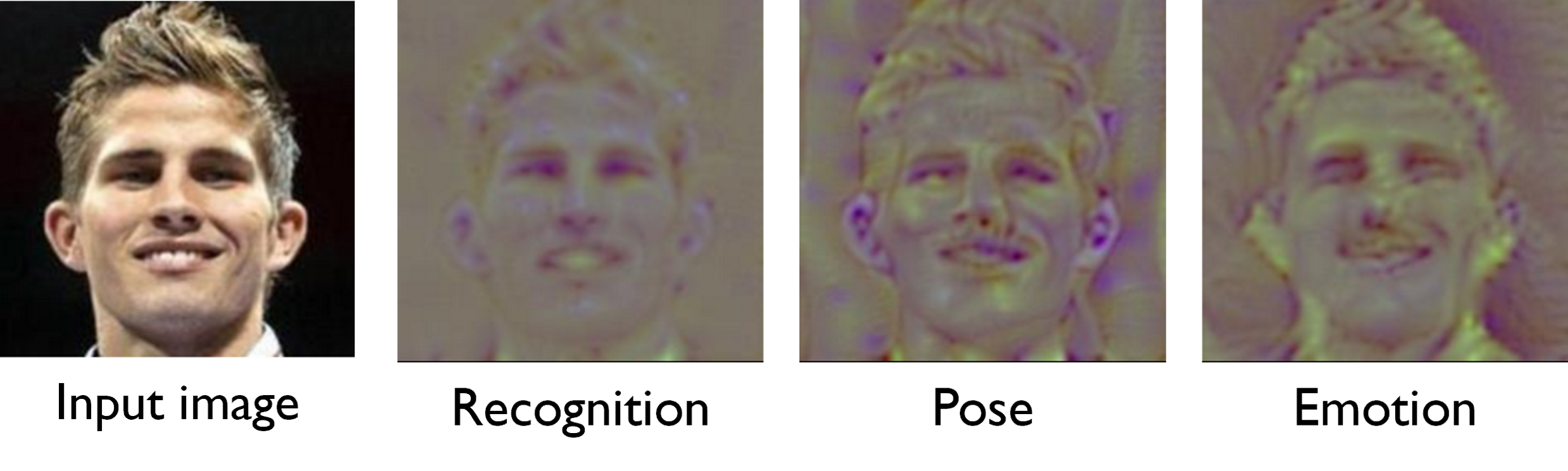}
	\vspace{-18pt}
	\caption{Result of inverting features of an input image from the Vgg-Face dataset \cite{parkhi2015vggface} using face networks trained for three different tasks. The models use VGG-16 architecture and are trained for recognition, pose and emotion. Inversions from the last convolutional layer from each network are shown here. We notice that the recognition network emphasizes the shape of the eyes and nose; pose network emphasizes the 3D shape; and the emotion network emphasizes the curves on the eyebrows and mouth.}
	\label{fig:face_inversion_networks}
\end{figure}
The results in Figure \ref{fig:face_inversion_methods} showed that irrespective of the feature inversion method, the face gets highlighted with some variations on the VGG-Face network. In Figure \ref{fig:face_inversion_networks}, we see the inversion of the same starting image from the features of face models from three different tasks. In each case, we observe that the network focuses on the discriminative information in the last convolutional layer. The recognition model focuses on the shape and proportions of facial features like the eyes and the nose. The head pose network does not faithfully reconstruct the shape of the features but retains the 3D information of the face. The emotion network exaggerates the curves in the face and eyebrows, as that information is crucial for detecting facial expressions. Such an observation may be useful when considering multi-task learning or meta-learning on face image data.

\begin{figure}
	\centering
	\includegraphics[width=\linewidth]{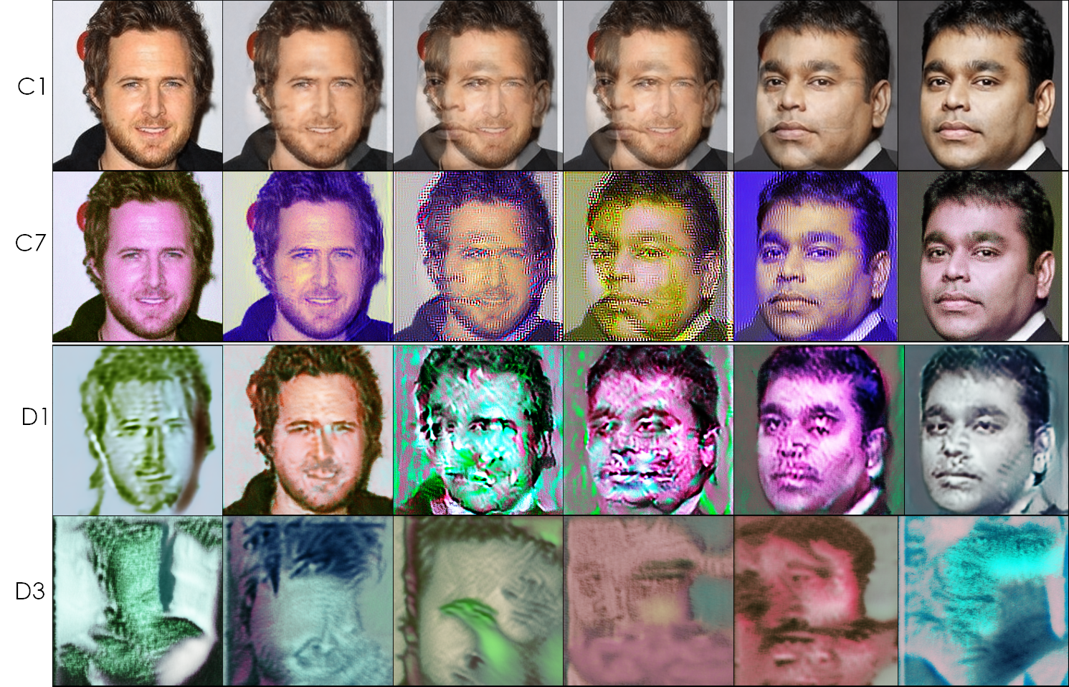}
	\caption{Interpolation in feature space between two face images. Features are extracted from a VGG-Face model trained with the Deep Image Prior \cite{ulyanov2018deep} regularizer. Original images are from the VGG-Face dataset \cite{parkhi2015vggface}.}
	\label{fig:interpolation_face}
\end{figure}

It is also helpful to visualize pre-images of feature vectors that do not correspond to a natural image. Figure \ref{fig:interpolation_face} shows the result of inverting features which are obtained by interpolation of features of two different face images. We extracted these features from various layers of the VGG-Face \cite{parkhi2015vggface} network. We see that interpolation in the convolutional layer features produces images that look like the two images added on top of each other. The dense layers create more semantic interpolation between the images.				
Feature inversion can be used in parallel with feature visualization to examine the hierarchy of deep features. Our experiments show that while the initial and middle layers can recreate the input image, the deeper layers focus on class-discriminative information related to the task being studied.
\section{Facial Concepts and Deep Face Features}
\label{sec:concepts}

The previous sections showed how different elements of a deep face model correspond to different facial features or parts of the face. For example, a neuron may respond to image patches that represent ‘nose’ or ‘left-facing’ or ‘rough skin’, but it is not immediately apparent by looking at a visualization of the neuron. This section investigates methods to align the meaning of these units to human-understandable concepts, usually represented as a textual description of the shared characteristics of a set of images. 

Should a concept correspond to a single neuron, or are random directions of neuron concepts equally meaningful? The answer depends on how `concept' is defined in a given setting. Szegedy et al. \cite{szegedy2013intriguing} hypothesized that random directions have as much semantic meaning as individual neurons. They collected images that produced a  high activation in individual units and compared them to images whose activations correspond to random directions of the units. Their experiments showed that both groups have semantic meaning. However, they used a loose definition of `concept', which described the common properties of a set of images. Whether such a concept would capture strong semantics may be an issue of concern. 
Other efforts \cite{bau2017networkdissection, zhou2018interpreting} leaned towards the view that neurons have special semantic meaning. They used a curated set of concepts comprising the Broden Dataset \cite{bau2017networkdissection } as their definition. They discovered that the natural basis, i.e. individual units, correspond to more unique concepts than orthogonal rotations of the basis with the same discriminative power as the natural basis. Again, this may be based on the limited dictionary of concepts provided by the Broden dataset. Fong et al. \cite{fong2018net2vec} contended that while random directions of neural units may not be as interpretable as the units, special directions exist which are more interpretable. They argued that the number of available feature channels is usually far smaller than the number of different concepts that a neural network may need to encode to interpret a complex visual scene. This suggests that, at the very least, the representation must use combinations of filter responses to represent concepts or, in other words, be at least in part distributed. They used the Broden dataset as their concept corpus, disregarding the scene and texture labels. Thus it is critical to define concepts correctly.

\begin{figure*}[t]
	\includegraphics[width = \linewidth]{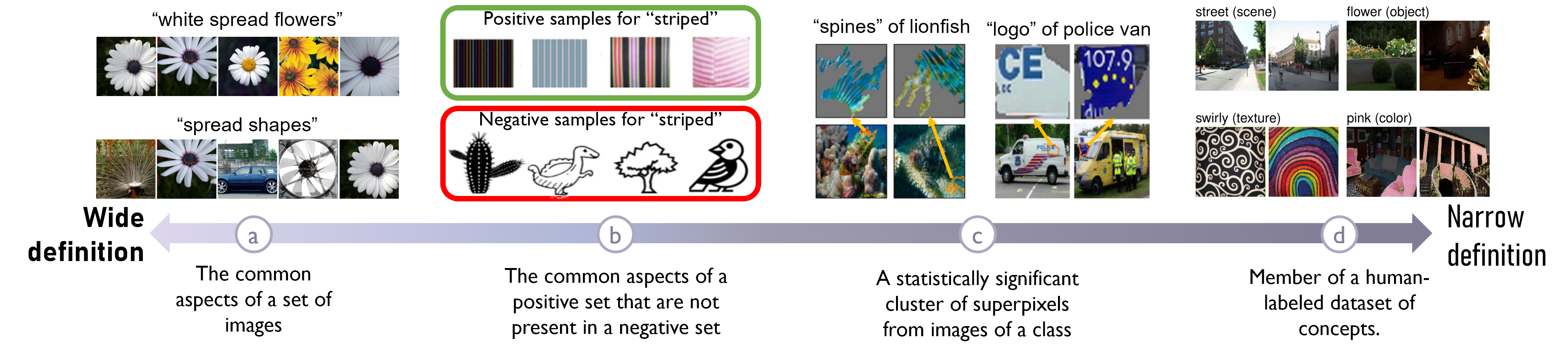}
	\caption{Different ways to define a `concept': (a) \cite{szegedy2013intriguing}, (b) TCAV \cite{kim2018tcav}, (c) ACE \cite{ghorbani2019towards}, (d) Broden \cite{bau2017networkdissection}.}
	\label{fig:concept_definition}
\end{figure*}

Can any shared property be a concept? Concepts are closely related to the categorizations of objects. This definition still allows room for arbitrary concepts, as we can find a categorization system for any random set of images. According to Rosch \cite{rosch1999principles}, human categorization should not be considered the arbitrary product of historical accident or whim but rather the result of psychological principles of categorization. Thus, some latent metric can make some concepts better than others. Here are some desired properties of concept definitions as described in \cite{ghorbani2019towards}: \textbf{Meaningfulness}: A concept is semantically meaningful on its own. Different individuals should associate similar meanings to the concept. \textbf{Coherency}: Examples of a concept should be perceptually similar to each other while being different from examples of other concepts. \textbf{Importance}: A concept is “important” for predicting a class if its presence is necessary to predict samples in that class accurately.

\begin{figure}
	\centering
	\includegraphics[width=\linewidth]{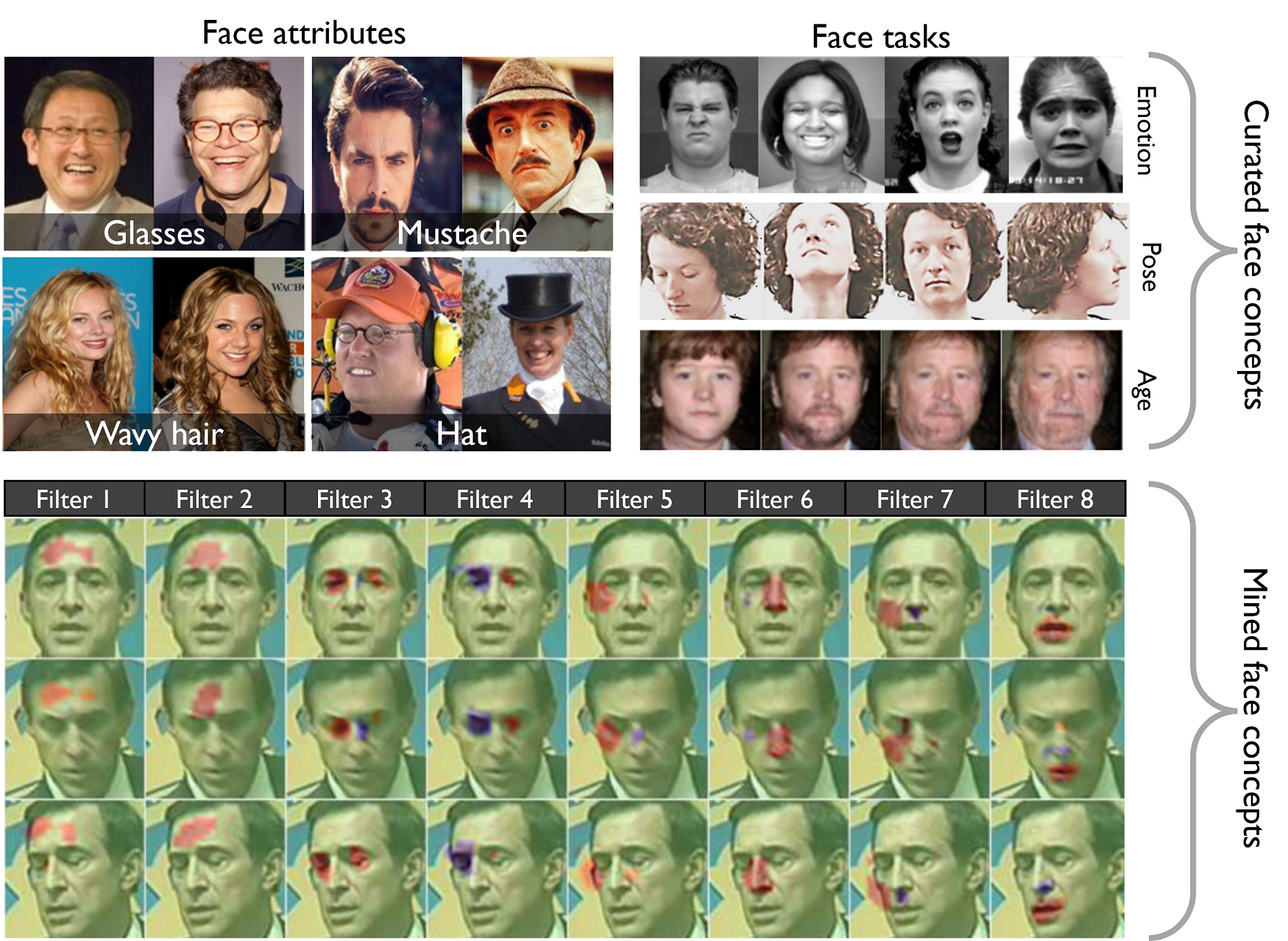}
	\caption{Listing of potential concepts for the face domain. \textit{Image sources:} Emotion: \cite{lucey2010extended}, Head pose: \cite{gourier2004estimating}, Age: \cite{antipov2017face}, Face attributes: \cite{liu2015CelebA}, Filter responses: \cite{yin2019Towards}.}
	\label{fig:face_concepts}
\end{figure}

Figure \ref{fig:concept_definition} shows the spectrum of concept definitions from weak to strong. We discuss two ways of defining concepts: curating a list of concepts or mining concepts from collections of images. Figs \ref{fig:concept_definition}(a) and \ref{fig:concept_definition}(c) represent concepts mined from a dataset. In case of  commonalities in a set of images \cite{szegedy2013intriguing} (Fig \ref{fig:concept_definition}(a)), concepts may include some amorphous concepts like "top round stroke” or “lower left loop” in the case of MNIST digits and “postures”, “spread shapes”, or “round green or yellow objects” for ImageNet images. Ghorbani et al. \cite{ ghorbani2019towards} introduced a more principled approach to automatically mining concepts from the image dataset called ‘Automatic Concept-based Evaluation’(ACE) (Fig \ref{fig:concept_definition}(c)). ACE searches for salient groups of superpixels in the dataset that satisfy some criteria for being a concept. They defined some tests for assessing concept definitions, based on how coherent humans find them. The first is the ‘Intruder Test’, which asks humans to pick out one semantically different image from a set of six images. The mined concepts are coherent if humans can pick the intruder consistently. In another test, humans are shown four concept segments and four random segments from images of the same class. They are asked to choose the most ‘meaningful’ grouping of image segments and describe their preferred option with one word. The number of individuals who use the same word to describe a group (after controlling for synonyms) measures the concept’s coherency. These metrics point to a ‘concept’ being a categorization that multiple individuals agree to. 

On the other hand, Figs \ref{fig:concept_definition}(b) and \ref{fig:concept_definition}(d) are concepts {curated by humans}. In particular, Kim et al. \cite{kim2018tcav} described a relaxed definition of a concept in TCAV (Fig \ref{fig:concept_definition}(b)), where each concept is defined by two image sets: one in which all images contain the concept and the other in which the concept is absent from all images. The Broadly and Densely Labeled (Broden) dataset \cite{ bau2017networkdissection} contains examples of a broad range of objects, scenes, object parts, textures, colors and materials in different contexts. Every class in Broden corresponds to an English word. 



The above concept-mining approaches have not been applied much to the face domain. Based on the above definitions and discussions, possible concepts for face image data include identity, pose, gender, race, facial hair and accessories. CelebA \cite{liu2015CelebA} is a dataset that has labels for 40 facial attributes, which could be considered concepts too. Facial parts, head pose, emotions and facial action units also constitute relevant concepts. Face concepts could also be defined based on the task of the pre-trained network, as in what may be concepts useful for classification or the specific task at hand.  In Figure \ref{fig:face_concepts}, we show possible definitions for facial concepts. Facial concepts may correspond to facial features like nose and eyebrows or be unique shapes or textures. 

In the only work in the face domain to the best of our knowledge, Yin et al. \cite{yin2019Towards} follow a different approach towards interpretability. Instead of ‘finding’ concepts corresponding to each convolutional filter, their modified training procedure pushes each filter to represent a concept.   Their approach uses a Siamese network with two branches sharing weights. The first branch gets a face image as input, and the second gets the same image superimposed with a synthetic occlusion. Along with the recognition loss, they introduce two new losses that encourage the filter representation to have a more consistent semantic meaning and require the filters to be insensitive to occlusions. The two losses ensure filter response locations are distributed across the face, and each filter concentrates on local face parts. Figure \ref{fig:face_concepts} (bottom) shows some filter responses from a model trained with this technique. We observe that each filter responds to a specific face feature regardless of the identity or pose of the face. This makes each filter’s representation more 'concrete' and more straightforward to describe in words, thus being easier for humans to conceptualize.

\section{Evaluation Protocols for Face Tasks}
\label{sec:faceevaluation}

Due to the fine-grained nature of the face domain, we need specialized evaluation protocols for face tasks. This section discusses creative evaluation methods that have been explicitly developed for faces. Instead of using aggregate statistics such as accuracy and ROC, RichardWebster et al. \cite{richardwebster2018psychophysics} suggested visual psychophysics as a viable methodology for making face recognition algorithms more explainable by the controlled manipulation of stimuli and careful study of the responses they invoke in a model system. This procedure is based on M-alternative forced-choice match-to-sample (M-AFC) psychophysics for general object classification \cite{webster2018psyphy}. The first step is 'herding' from an initial dataset, where we remove identities that consistently cause false matches or false non-matches, i.e. errors that are inherent within the matching system. This step removes the `\textit{goats}', `\textit{lambs}', and `\textit{wolves}' from the `\textit{sheep}', where the \textit{goats} are challenging to match, \textit{lambs} are identities that are easy to impersonate, and \textit{wolves} easily impersonate other identities. \textit{Sheep} are well-behaved identities that match well to themselves but poorly to others. In the second step, we create item-response curves' by gradually perturbing the images and recording the match rate. Possible perturbations include noise, contrast, blur, blink, and expression. Results from psychophysics experiments using highly controlled procedurally generated stimuli can then inform how we should use a face recognition algorithm by explaining its failure modes.

\section{User Survey: Utility of Face Explainability}
\label{sec:usersurvey}

The previous sections discussed various explainability algorithms developed over the years pursuing goals of correctness and clarity. In practice, there may be a domain gap between practitioners and consumers of explainable deep learning methods. Here, `practitioners' develop and improve explainability algorithms, whereas `consumers’ use algorithms to make decisions on the correctness, viability, and trustworthiness of their models. The consumers may not have in-depth knowledge on the working of these algorithms and hence may not be equipped to make the same conclusions about models as the practitioners. We conducted a user study to determine how helpful popular explainability methods are in answering questions about the decisions of deep learning-based face models. 

\begin{figure}
	\includegraphics[width=\linewidth]{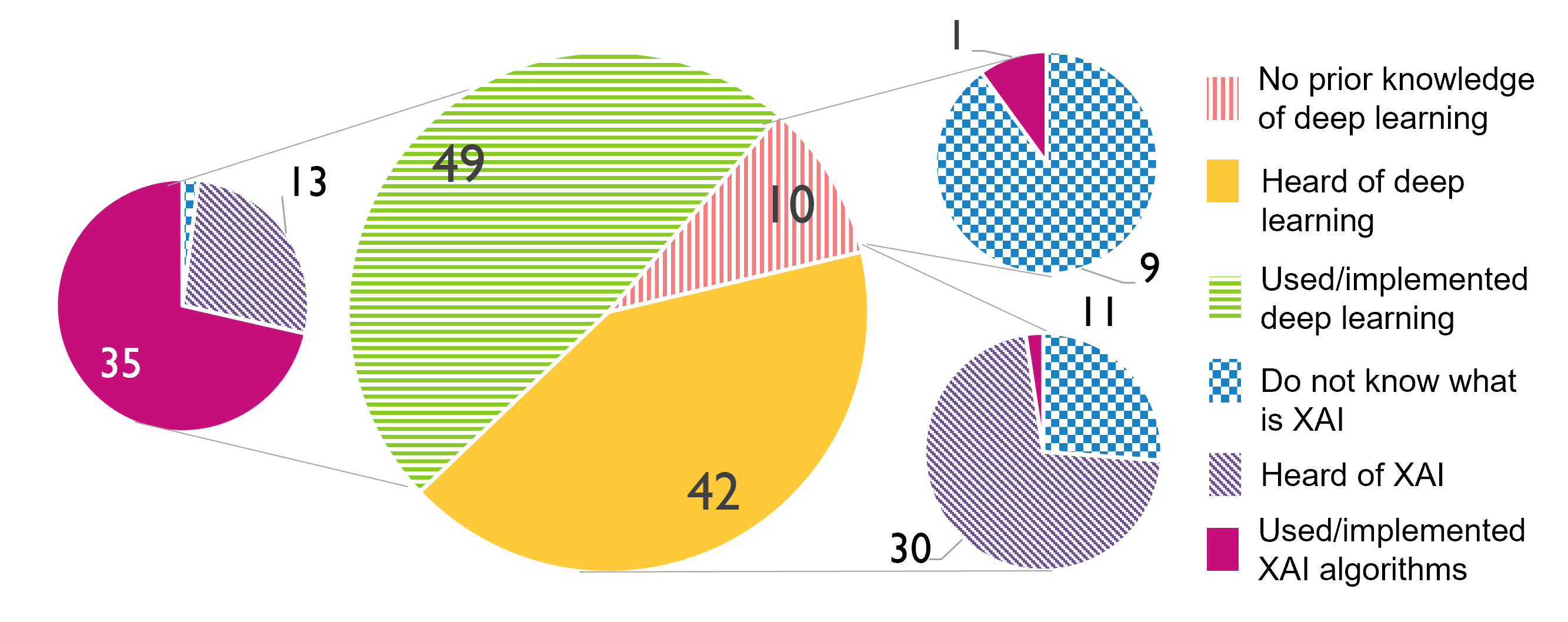}
	\caption{Demographic of participants who responded to the survey in our user study.}
	\label{fig:userstudydemo}
\end{figure}

\begin{figure}
	\includegraphics[width=\linewidth]{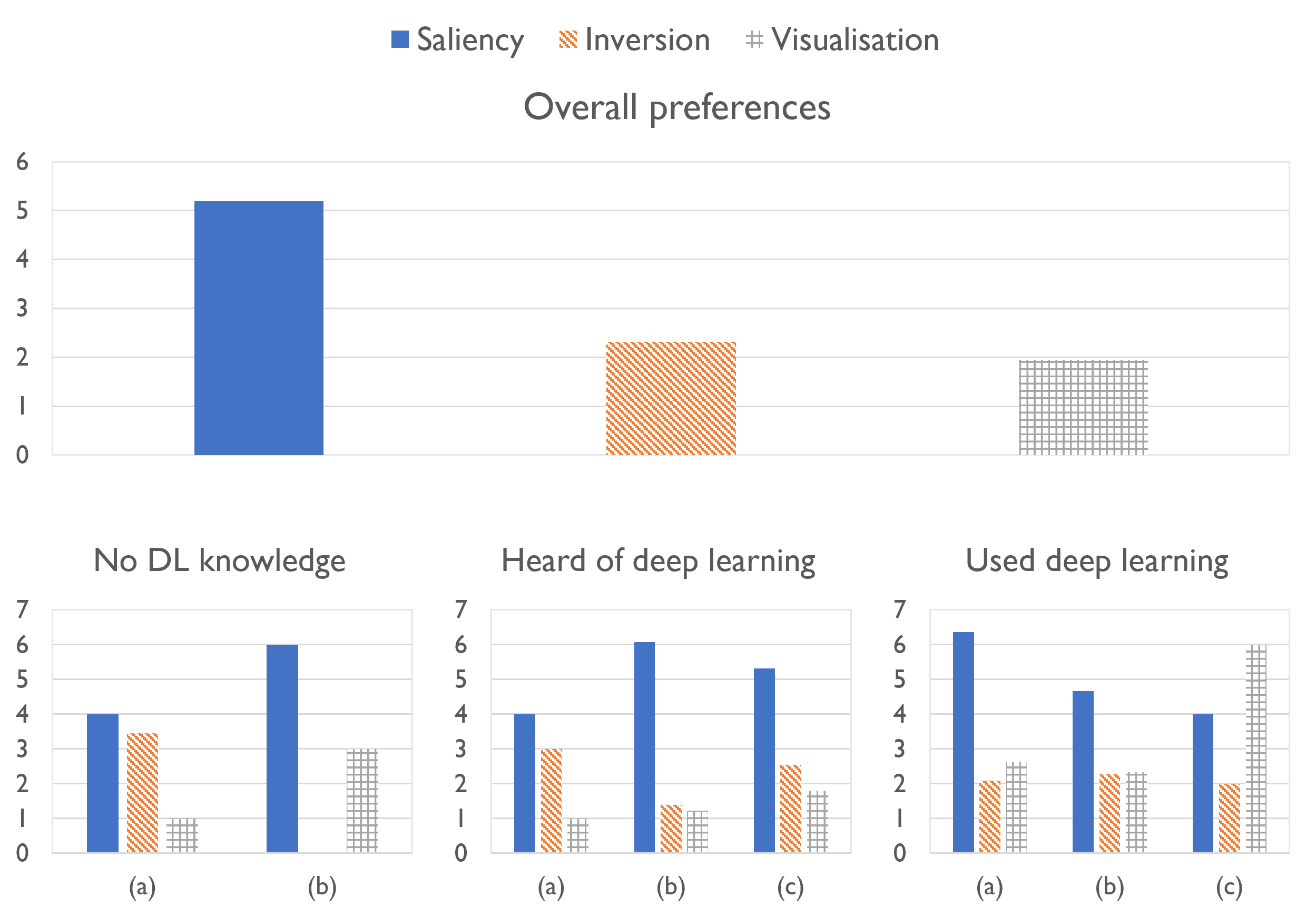}
	\caption{Results of our user study. Top row shows the overall aggregated preference, while the bottom row shows preferences categorized by their familiarity with deep learning. The results correspond to: (a) participants who have never heard of explainability, (b) participants who understand the need for explainability but have never used it, and (c) participants who have used or implemented explainability algorithms.}
	\label{fig:userstudyresults}
\end{figure}

In particular, we chose three popular explainability strategies: (a) saliency map, (b) feature inversion, and (c) feature visualization. We computed saliency maps using occlusion \cite{zeiler2014visualizing}, while we used total variation regularization \cite{mahendran2015inverting} for both feature inversion and feature visualization. 
We studied VGG16 models trained for gender, age and expression and generated explanations for eight input images. We presented eight questions to each respondent and asked each of them to pick the explanation(s) that best helped them to understand a model prediction. Sample questions included: "Which explanation best helps you respond to the question, ``Why is this person classified to be in his 40s?'', and ``Which of these explanations best helps you see why this person has an angry expression?''.
Of 101 participants in our study, 48.5\% were deep learning practitioners, and 35.6\% had used or implemented explainability methods (Figure \ref{fig:userstudydemo}). As shown in Figure \ref{fig:userstudyresults}, most respondents preferred saliency maps, except those who have used or implemented explainability algorithms. In most cases, people preferred feature inversion over feature visualization. The group that consisted of people who had implemented both deep learning and explainability algorithms preferred feature visualization over saliency maps. 

These results suggest a preference gap between practitioners and consumers of explainability algorithms. This may be because saliency maps are more useful for an overall view of the model, whereas feature visualization gives an internal view useful to developers of the model. Saliency maps may also be easier to interpret than feature visualization and feature inversion. Our study indicates the need to consider usability and include end consumers in the development process of explainability methods targeted for specific face processing tasks.

\section{Conclusion}
In this work, we study visual explanation/explainability methods with a view of their application to facial image processing and understanding. We present a summary of different categories of explainability methods relevant to this objective, and also discuss specialized face explanation algorithms and evaluation protocols designed for face explainability algorithms. 
By applying these methods to popular deep learning-based face models, we compare these methods and identify subtle differences between the methods that can be relevant and useful in face processing tasks. We also presented effective design criteria for the adoption of explainability algorithms, especially in face processing tasks.

A key takeaway from our study is that explainability methods on the face domain may need to have their own considerations, and not necessarily align with more general uses of explainability algorithms. However, despite the extensive work on explainability methods for deep learning, only a few are tailored toward faces. We need more efforts in this direction, especially considering face processing models are used in safety-critical applications such as biometrics. Besides, our user study indicated that end users need to be involved in the design and deployment of such models. We hope that the detailed overview, experimental studies and analyses presented in this work will lead to wider and well-founded adoption of explainability methods in face processing tasks. 

\bibliographystyle{IEEEtran}
\bibliography{IEEEabrv,references}

\begin{IEEEbiography}[{\includegraphics[width=1in,height=1.5in,clip,keepaspectratio]{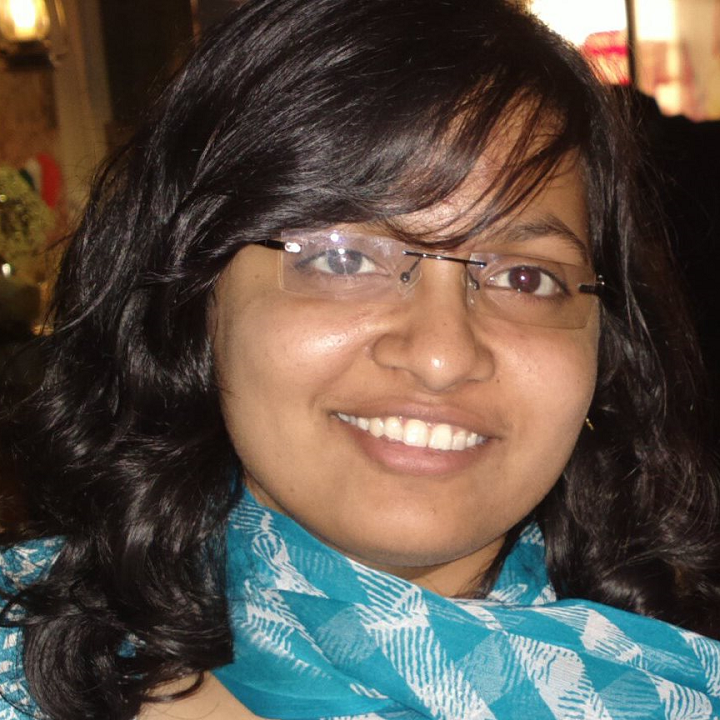}}]{Thrupthi Ann John} recieved the B.Tech in Computer Science from the National Institute of Technology, Warangal. She is currently a PhD student in the Computer Vision and Information Technology lab in IIIT Hyderabad. She is advised by Prof C V Jawahar and Prof Vineeth N Balasubramanian. She has published in top venues including TBIOM and IROS. She is a recipient of the Visvesvaraya PhD Fellowship from Ministry of Electronics and Information Technology, Government of India. Her interests include deep algorithms for face tasks and explainability for deep learning. 
\end{IEEEbiography}

\begin{IEEEbiography}
	[
	{\includegraphics[width=1in,height=1.5in,clip,keepaspectratio]{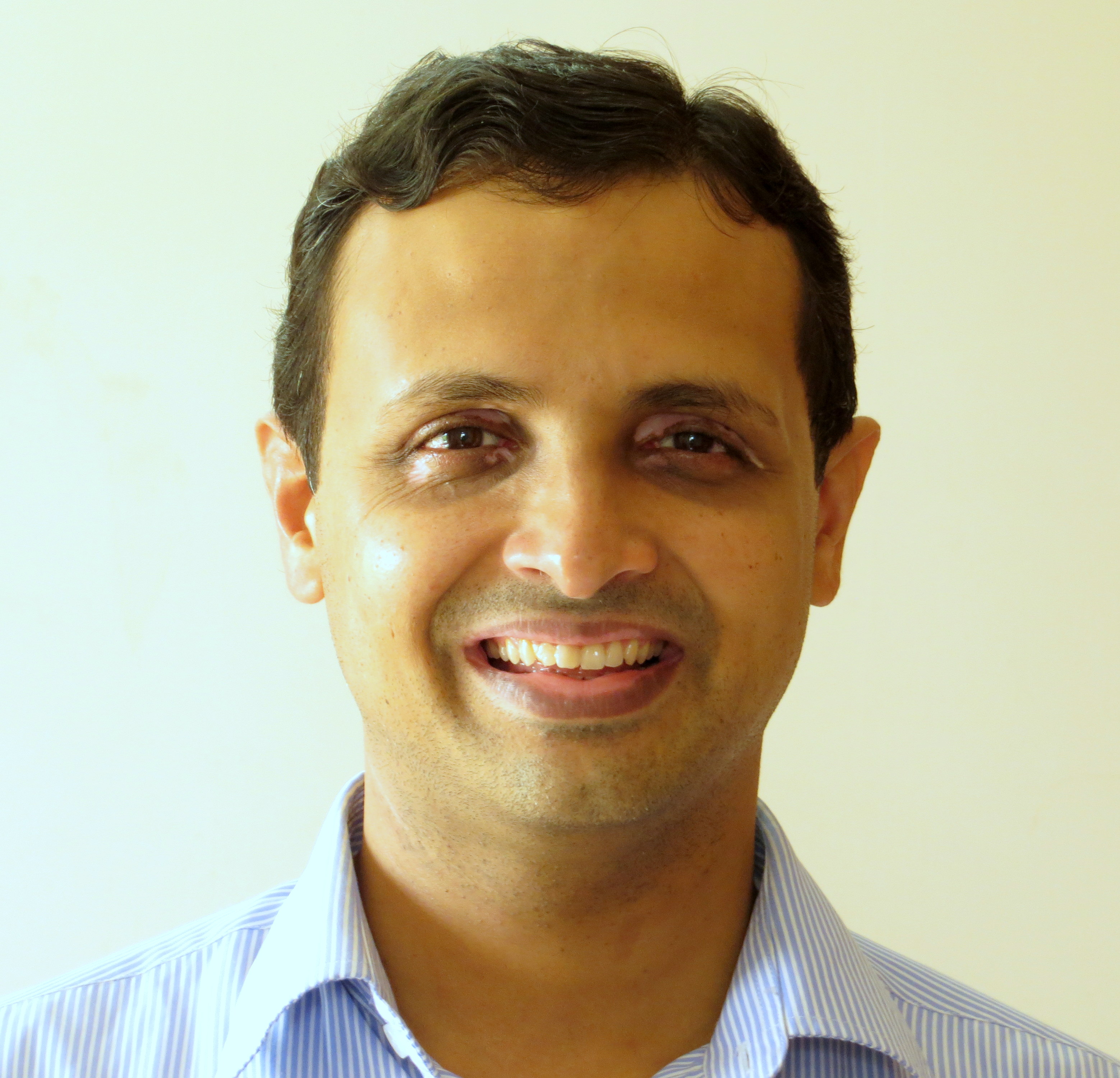}}
	]%
	{Vineeth N Balasubramanian}
	is an Associate Professor in the Department of Computer Science and Engineering at the Indian Institute of Technology, Hyderabad (IIT-H). His research interests include deep learning, machine learning, and computer vision, with a focus on explainable deep learning and learning with limited supervision. His research has resulted in over 100 peer-reviewed publications at various international venues, including top-tier venues such as ICML, CVPR, NeurIPS, ICCV, KDD, ICDM and IEEE TPAMI. He regularly serves as a senior PC/Area Chairfor conferences such as CVPR, ICCV, AAAI and IJCAI, with recent awards as Outstanding Reviewer at ICLR 2021, CVPR 2019, ECCV 2020 etc. He is also a recipient of the Google Research Scholar Award 2020-21.  For more details, please see \url{https://iith.ac.in/~vineethnb/}.
\end{IEEEbiography}\

\begin{IEEEbiography}[{\includegraphics[width=1in,height=1.5in,clip,keepaspectratio]{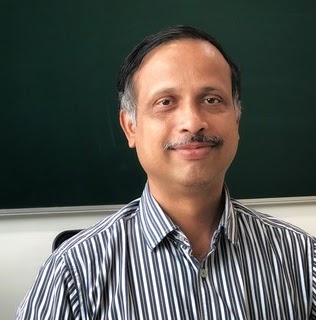}}]{C V Jawahar}
is the Dean of Research and Development and Head of the Centre for Visual Information Technology-CVIT, at the International Institute of Information Technology, Hyderabad (IIITH), India. He leads the research group focusing on computer vision, machine learning, and multimedia systems. He was awarded a doctorate from IIT Kharagpur. He is  an elected Fellow of INAE and IAPR. His prolific research is globally recognized in the Artificial Intelligence and Computer Vision research community with more than 100 publications in top tier conferences and journals in computer vision, robotics and document image processing to his credit with over 12000 citations. He is awarded the ACM India Outstanding Contribution to Computing Education (OCCE) 2021. He is actively engaged with several government agencies, ministries, and leading companies around innovating at scale through research.
\end{IEEEbiography}

\ifCLASSOPTIONcaptionsoff
  \newpage
\fi

\end{document}